\documentclass[twoside,twocolumn,9pt]{article}
\usepackage{array}
\usepackage{fancyvrb}
\usepackage{listings}
\usepackage{booktabs}
\usepackage{makecell}
\usepackage{tabularx}
\usepackage{array} 

\usepackage{amsmath,amssymb}

\usepackage{extsizes}
\usepackage[super,sort&compress,comma]{natbib} 
\usepackage[version=3]{mhchem}
\usepackage[left=1.5cm, right=1.5cm, top=1.785cm, bottom=2.0cm]{geometry}
\usepackage{balance}
\usepackage{mathptmx}
\usepackage{sectsty}
\usepackage{graphicx} 
\usepackage{lastpage}
\usepackage[format=plain,justification=justified,singlelinecheck=false,font={stretch=1.125,small,sf},labelfont=bf,labelsep=space]{caption}
\usepackage{float}
\usepackage{fancyhdr}
\usepackage{fnpos}
\usepackage[english]{babel}
\addto{\captionsenglish}{%
  
}
\usepackage{array}
\usepackage{droidsans}
\usepackage{charter}
\usepackage[T1]{fontenc}
\usepackage[usenames,dvipsnames]{xcolor}
\usepackage{setspace}
\usepackage[compact]{titlesec}
\usepackage{hyperref}

\usepackage{epstopdf}

\definecolor{cream}{RGB}{222,217,201}

\begin{document}

\twocolumn[
\begin{@twocolumnfalse}

\title{\textbf{DrugR: Optimizing Molecular Drugs through LLM-based Explicit Reasoning}}

\author{%
Haoran Liu$^{1}$,
Zheni Zeng$^{2}$\thanks{Corresponding author},
Yukun Yan$^{3}$,
Yuxuan Chen$^{4}$,
and Yunduo Xiao$^{5}$
}

\date{} 

\maketitle

{\normalsize\noindent\textbf{Abstract.}~
Molecule generation and optimization is a fundamental task in chemical domain.
The rapid development of intelligent tools, especially large language models (LLMs) with powerful knowledge reserves and interactive capabilities, has provided new paradigms for it.
Nevertheless, the intrinsic challenge for LLMs lies in the complex implicit relationship between molecular structure and pharmacological properties and the lack of corresponding labeled data.
To bridge this gap, we propose \textbf{DrugR}, an LLM-based method that introduces explicit, step-by-step pharmacological reasoning into the optimization process.
Our approach integrates domain-specific continual pretraining, supervised fine-tuning via reverse data engineering, and self-balanced multi-granular reinforcement learning.
This framework enables DrugR to effectively improve key ADMET properties while preserving the original molecule's core efficacy.
Experimental results demonstrate that DrugR achieves comprehensive enhancement across multiple properties without compromising structural similarity or target binding affinity.
Importantly, its explicit reasoning process provides clear, interpretable rationales for each optimization step, yielding actionable design insights and advancing toward automated, knowledge-driven scientific discovery.
Our code and model checkpoints are open-sourced to foster future research.\par}

\vspace{1em} 

\end{@twocolumnfalse}
]

\renewcommand*\rmdefault{bch}\normalfont\upshape
\rmfamily
\titlespacing*{\section}{0pt}{0pt}{0pt}
\vspace{-1cm}

\begingroup
\renewcommand{\thefootnote}{} 

\footnotetext{\textit{$^{1}$~School of Biological Science and Medical Engineering, Beihang Advanced Innovation Center for Biomedical Engineering, Beihang University, Beijing 100191, China.}}
\footnotetext{\textit{$^{2}$~Nanjing University.}}
\footnotetext{\textit{$^{3}$~Tsinghua University.}}
\footnotetext{\textit{$^{4}$~School of Electronic and Computer Engineering, Peking University, Shenzhen, China.}}
\footnotetext{\textit{$^{5}$~School of Computer Science and Engineering, South China University of Technology, Guangzhou, China.}}

\addtocounter{footnote}{-5} 
\endgroup



\section{Introduction}

Molecule optimization is a crucial step in drug discovery, involving the partial modification of molecules to improve the expected properties of candidate drugs. Machine learning-driven computational systems can help reduce reliance on human expert experience to some extent, but significant challenges remain in practical applications: (1) Complexity: How to perform multi-objective optimization? (2) Generalization: How to adapt to various types of drugs and their properties? (3) Interpretability: How to provide clear and reliable decision-making criteria?

With the appearance of LLMs, we have discovered new possible solutions to the above problems. Compared with smaller models, LLMs can follow flexible instructions to optimize multiple properties, and may adapt to different subdomains with their extensive knowledge accumulation. They also demonstrate excellent reasoning abilities, such as the impressive deep thinking function of models like DeepSeek-R1~\cite{guo2025deepseek}. Theoretically, by simply providing the properties of the original molecule to an LLM with prompts, a new molecular structure comprehensively optimized can be derived through reasoning. However, in practice, we have found that general-domain LLMs lack knowledge of molecular structure and drug properties, while relatively small-scale domain models suffer significantly impaired reasoning capabilities and cannot directly perform this task effectively. Therefore, we decide to design a model training method that enables explicit reasoning-driven optimization of drug molecules, and name it DrugR.

\textbf{Task Definition}. To facilitate molecular representation and property prediction, we limit the optimization scope to three classes of small molecule drugs: anti-inflammatory, antihypertensive, and hypoglycemic. Provided with the original drug molecule and its corresponding pharmacological properties, the system is required to generate a new molecule (in SMILES expression) that comprehensively optimizes those poor properties through reasoning. Considering the structural and functional consistency between the two molecules, the optimized result should have a fingerprint similarity with the original one higher than 0.6.
To conduct automatic efficient evaluation, a property simulator has to be introduced. In this work, we adopt the ADMETLab~\cite{admetlab3} tool to quickly estimate 23 types of molecular pharmacological properties (including QED, permeability, toxicity and so on). 

\begin{figure}[t]
  \centering
  \includegraphics[width=\columnwidth]{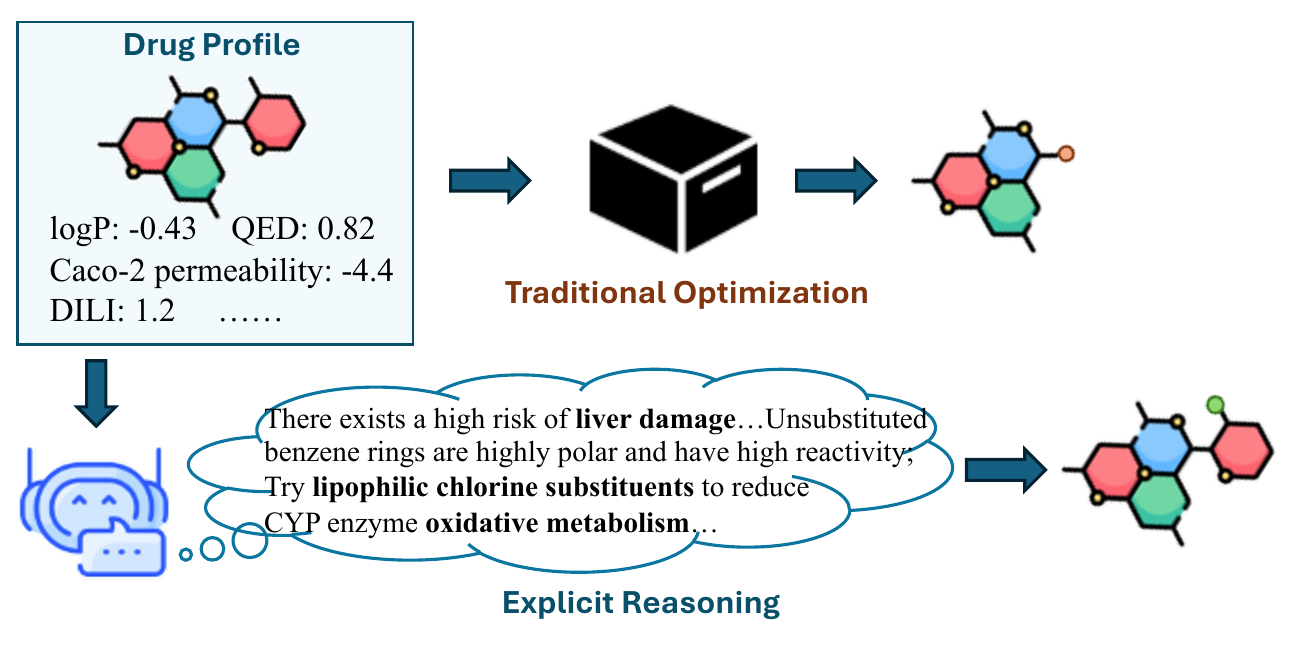}
  \caption{Schematic for explicit reasoning-based molecule optimization.}
  \label{fgr:schematic}
\end{figure}

\textbf{Data Construction}. Since there is neither readily available molecule optimization reasoning data, nor known systems that can perform this task, we combine the capabilities of professional tools, strong LLMs and domain knowledge bases to conduct reverse data engineering. Over 10,000 existing small-molecular drugs are collected from DrugBank\footnote{\url{https://go.drugbank.com/}} as seed data, of which a group of similar structures are randomly generated and evaluated by the simulator. Within the same group, those molecules having better pharmacological properties are marked as positive items, and worse as negative. With the pairwise data, Deepseek-R1 is utilized to annotate the reasoning process from negative to positive molecules, which is a much easier task than directly designing positive ones.

\textbf{Model Training}. We adopt LLaMA-3-8B-Instruct\cite{grattafiori2024llama3} as our backbone model, which is an edge LLM with relatively comprehensive performance, suitable for validating our method. The training process is divided into three stage. (1) \textit{Continual pre-training} (CPT): we collect domain corpus like knowledge bases and literature records to help enhance the chemical knowledge of the model. Both general and domain instruction tuning data is used to prevent catastrophic forgetting. (2) \textit{Supervised fine-tuning} (SFT): we use the task data constructed by reverse data engineering to train the capability of recognizing bad properties and designing better structures. (3) \textit{Reinforcement learning} (RL): we conduct the Group Relative Policy Optimization (GRPO)~\cite{shao2024deepseekmath} algorithm with the multi-granular reward function. Different fields of the model output are assigned different reward content, including feature localization accuracy, reasoning diversity, design effectiveness, drug efficacy retention and so on. We also find Pareto improvement strategy useful to achieve a self-balance between multiple reward issues in this stage.

To evaluate the performance of DrugR, we compare the designing results of our method with the backbone and other strong baselines using computational tools such as ADMETLab and a docking-based binding estimation pipeline. It is demonstrated that DrugR can obviously optimize the key pharmacological scores (relative increase of 89.5\%) while still keep a high fingerprint similarity with the original input, and therefore maintain a relatively good target binding affinity (only worsened by about 4.1\%). Ablation study shows the effectiveness and generalization of our strategies, and case analysis proves the professionalism and innovation of the LLM reasoning process as its decision-making basis. This will also be an interesting routine for hidden knowledge extraction.

We open-source our code, model, and dataset to advance research in reasoning-driven drug discovery. More importantly, we advocate for the integration of such systems into real-world drug development pipelines. This would create a closed-loop, data-driven flywheel: experimental results (including failed trials) continuously refine the model, which in turn generates higher-quality candidates for subsequent testing, ultimately accelerating the entire discovery process.

\section{Results}

\subsection{Evaluation Metrics}

We evaluate molecular optimization from a multi-dimensional perspective along
three complementary axes: (1) Pharmacological property improvement, which is estimated according to the comprehensive property results predicted by the simulator; (2) 
Functional and structural consistency, represented by binding affinity and molecular fingerprint similarity; (3) Explicit reasoning validity, automatically evaluated by LLMs.
These metrics work together to capture not only the quality of optimized molecules,
but also the faithfulness and interpretability of the optimization process. Here we provide a brief description of the principles behind each metric. Details are provided in section~\ref{sec:method}.

\paragraph*{Pharmacological Property Improvement.}

Molecular optimization aims to improve the pharmacological profile of a candidate subject to feasibility constraints rather than to satisfy a fixed set of target values. Consequently, the same numerical change may be beneficial or detrimental depending on the pharmacological context and the induced trade-offs. In our setting, 23 indicators for specific properties are comprehensively considered, and different normalization functions are designed for range-targeted, threshold-based, and free indicators. We eventually define an overall optimization score from 0 to 1, which is the weighted sum for the direction-aware score of each item. If the optimized molecule has overcome all unsatisfying properties of the original one, and does not appear new shortcomes, then the overall score will be 1. In contrast, when none of the target properties is improved, it will be 0.

\paragraph*{Functional and Structural Consistency.}
While improving pharmacological properties, realistic lead optimization requires
preserving the core molecular scaffold and target-related functionality of the
original compound. Fingerprint similarity is used to quantify structural preservation between
the original and optimized molecules.
We compute the Tanimoto similarity based on extended-connectivity fingerprints
(ECFP4), a widely adopted representation for capturing local chemical environments
in medicinal chemistry.
A higher similarity score indicates that pharmacological improvements are achieved
with minimal deviation from the original chemical scaffold, which aligns with
practical lead optimization workflows~\cite{rogers2010ecfp, bajusz2015tanimoto}.

Meanwhile, docking-based binding scores serve as a functional constraint for checking whether an optimized molecule retains target-level interactions consistent with the original mechanism of action~\cite{meng2011docking,ferreira2015docking}. The score is not interpreted as a direct surrogate for clinical efficacy; instead, it offers a mechanistically grounded and computationally practical indicator of functional preservation that is widely used in structure-based lead optimization~\cite{lu2010residencetime}.

Considering the nature of drug molecule optimization tasks, we emphasize gains in the overall optimization score as the primary objective, while requiring both binding affinity and fingerprint similarity to remain within acceptable ranges. Specifically, we define a similarity higher than 0.6 and binding energy lower than -6 kcal/mol as reasonable. This design reflects practical lead optimization scenarios, where global pharmacological improvement is prioritized without compromising target-level functionality or core chemical structure.

\paragraph*{Reasoning Validity.}
Since we expect our method to explicitly generate optimization rationales, we evaluate not only what molecular modifications are proposed, but also whether the rationale correctly identifies the ADMET liabilities that motivate optimization. 
We first quantify this diagnostic accuracy using Target Property F1, which measures feature-level agreement between the liabilities mentioned in the reasoning process, and the simulator-predicted problematic properties of the input molecule.

As for more comprehensive evaluation, since the reasonable optimization rationales are not unique, we can not directly calculate the superficial textual similarity between the reference answer and the generated response to represent its quality. Thus we introduce Language Modeling Score (LMS) produced by an automatic judge (gpt-4o-mini) to reflect the strategy-level soundness \cite{openai_gpt4omini_docs}. Specifically, LMS is derived from three complementary considerations: alignment between the stated rationale and the realized molecular edit (evidenced by SMILES differences and accompanying property changes), scientific and chemical validity without substantive factual or causal mistakes, and the plausibility of the implied optimization strategy. We further incorporate a reasoning-richness evaluation, which favors informative variation without sacrificing grounding.
It is worth noting that these evaluation metrics also serve as rewards in the reinforcement learning stage of our method. More detailed discussions on them are provided in Section~\ref{section:reward}.

\subsection{Optimization Results}

\begin{table*}[t]
\centering
\caption{Overall optimization performance comparison across baselines and our approach. {\color{red}Red} stands for unreasonable average similarity.}
\label{tab:main_results}
\small
\setlength{\tabcolsep}{6pt}
\begin{tabular}{lccccc}
\toprule
\textbf{Baseline} &
\textbf{Overall Optimization Score} &
\textbf{Target property F1 score} &
\textbf{Fingerprint Similarity} &
\textbf{reasoning LMS score} &
\textbf{reasoning richness} \\
\midrule
diffusion       & 0.1326 & --      & 0.8083 & --         & -- \\
mobo            & 0.0610 & --  & 0.9992 & --         & -- \\
\midrule
GPT5            & 0.1969 & 0.2209  & {\color{red}0.3677} & 0.5721         & 0.8570 \\
DeepSeek-R1     & 0.1787 & 0.1911  & {\color{red}0.4793} &  0.7011         & 0.8462 \\
\midrule
ChemDFM         & 0.1997 & 0.1326  & {\color{red}0.1474} & 0.5272         &  0.3165  \\
ExLLM           & 0.1429 & 0.1154  & 0.7291 & 0.5270         & 0.4613 \\
ether0          & 0.2149 & 0.1265  & {\color{red}0.2207} & 0.6116        & 0.4613 \\
LLaMA3-8B       & 0.1551 & 0.1354  & {\color{red}0.2843} &  0.5339       &  0.7411  \\
\midrule
\textbf{DrugR} & \textbf{0.2712} & \textbf{0.4364} & 0.6409 & \textbf{0.7712} &  \textbf{0.9877} \\
\ \ \textit{w/o RL} & 0.2330 & 0.3155 & 0.7809 &  0.7196   & 0.9846 \\
\ \ \textit{Pre-training} & 0.0836 & 0.0773 & {\color{red}0.2667} & 0.3774         & 0.0073 \\
\ \ \textit{SFT}& 0.2293 & 0.4290  & 0.7992 & 0.6874     &  0.9795 \\
\ \ \textit{w/o reasoning}    & 0.0151 & --      & {\color{red}0.1297} & --         & -- \\
\bottomrule
\end{tabular}
\end{table*}

\begin{table*}[t]
\centering
\caption{Out-of-Distribution (OOD) adaptation results on the anticancer dataset. \textsuperscript{*} indicates that \textbf{DrugR} is fine-tuned on the small amount of new data before evaluation.}
\label{tab:ood}
\small
\setlength{\tabcolsep}{6pt}
\begin{tabular}{lccccc}
\toprule
\textbf{Baseline} &
\textbf{Overall Optimization Score} &
\textbf{Target property F1 score} &
\textbf{Fingerprint Similarity} &
\textbf{reasoning LMS score} &
\textbf{reasoning richness} \\
\midrule
ether0          & 0.1076 & 0.0586  & \color{red}0.2580       & 0.6434   & 0.4628 \\
LLaMA3-8B       & 0.1594 & 0.0368    & \color{red}0.3771  &  0.4996      &  0.3540  \\
ExLLM       & 0.1453 & 0.0385    & \color{red}0.4819  &  0.6540     &  0.4875  \\
\midrule
\textbf{DrugR}$^{*}$ 
& \textbf{0.2060} & \textbf{0.3404}  & \textbf{0.6327}  & \textbf{0.6693} &  \textbf{0.5210} \\
\ \ \textit{SFT}$^{*}$& 0.1949 & 0.2997  & \color{red}0.5380  & 0.6051  &  0.4520 \\
\bottomrule
\end{tabular}
\end{table*}

We present the main optimization results by comparing our method against a
diverse set of baselines that span (i) classical non-LLMs molecular optimization
methods, (ii) general-purpose LLMs, (iii) chemistry-adapted molecular
reasoning/generation models, and (iv) ablations of our framework.
The selected baselines are chosen to reflect both classical optimization paradigms
and representative LLM-based reasoning systems. Notice that the LLM-based systems read and generate molecule structures with SMILES strings, and others directly process molecule comformations. In particular, DeepSeek-R1 is included as a key baseline since it serves as the annotator
model for generating reference rationales, making it a strong and well-aligned baseline for reasoning-guided optimization. We additionally evaluate general LLMs with varying capacities to assess
how model scale and domain knowledge affect performance on our tasks.

As shown in Table~\ref{tab:main_results}, our method achieves the best performance on the primary optimization metric, significantly outperforming all baselines in terms of overall optimization score, while maintaining reasonable fingerprint similarity to the original molecules. Relative to the annotated molecules in our dataset (Overall Optimization Score: 0.1653; Target-property F1: 0.3704), DrugR even increases the overall optimization score by 63.3\% and improves inference F1 by 17.8\%. In our observation, almost all methods that have not been specifically trained on the target task data exhibit low similarity scores (defined as less than 0.6). This includes the powerful general-purpose model GPT-5 and DeepSeek-R1, the chemistry-domain models ChemDFM\cite{zhao2024chemdfm} and ether0~\cite{narayanan2025ether0}, and the backbone model LLaMA3-8B-Instruct. They fail to adhere to the given molecular precursor structure and frequently produce results similar to existing common drugs, making them unsuitable for real-world drug design processes. Therefore, their overall optimization score may be high while meaningless.

In contrast, traditional generative approaches such as diffusion-based~\cite{ChangYe2024LDMol} and mobo-based systems~\cite{Zhu2023SampleEfficientMO}, can better adhere to the given original molecular structure and make minor adjustments, but the improvement in drug properties is not significant. In particular, we observe that diffusion model tends to optimize averaged objectives and lacks fine-grained control over property-specific trade-offs, leading to suboptimal improvements and unstable convergence when objectives conflict. Note that these methods do not generate explicit reasoning, thus the target property F1 and reasoning scores are omitted.

Among all these baselines, ExLLM~\citep{ran2025exllm} learns plenty of molecular-specific knowledge during its pre-training, and maintains its general capabilities to some extent. Therefore, it can basically correctly understand and execute pharmacology optimization tasks. Nevertheless, the low scores in terms of target property F1 and LMS show its poor reasoning quality. As for GPT-5 and DeepSeek-R1, though with rich knowledge and strong reasoning capability, it remains difficult to precisely constrain the behavior of the task through prompting engineering, which illustrates the challenge of comprehensive optimization of molecular pharmacology.

\subsection{Binding Affinity Analysis}

\begin{figure}[t]
    \centering
    \includegraphics[width=\columnwidth]{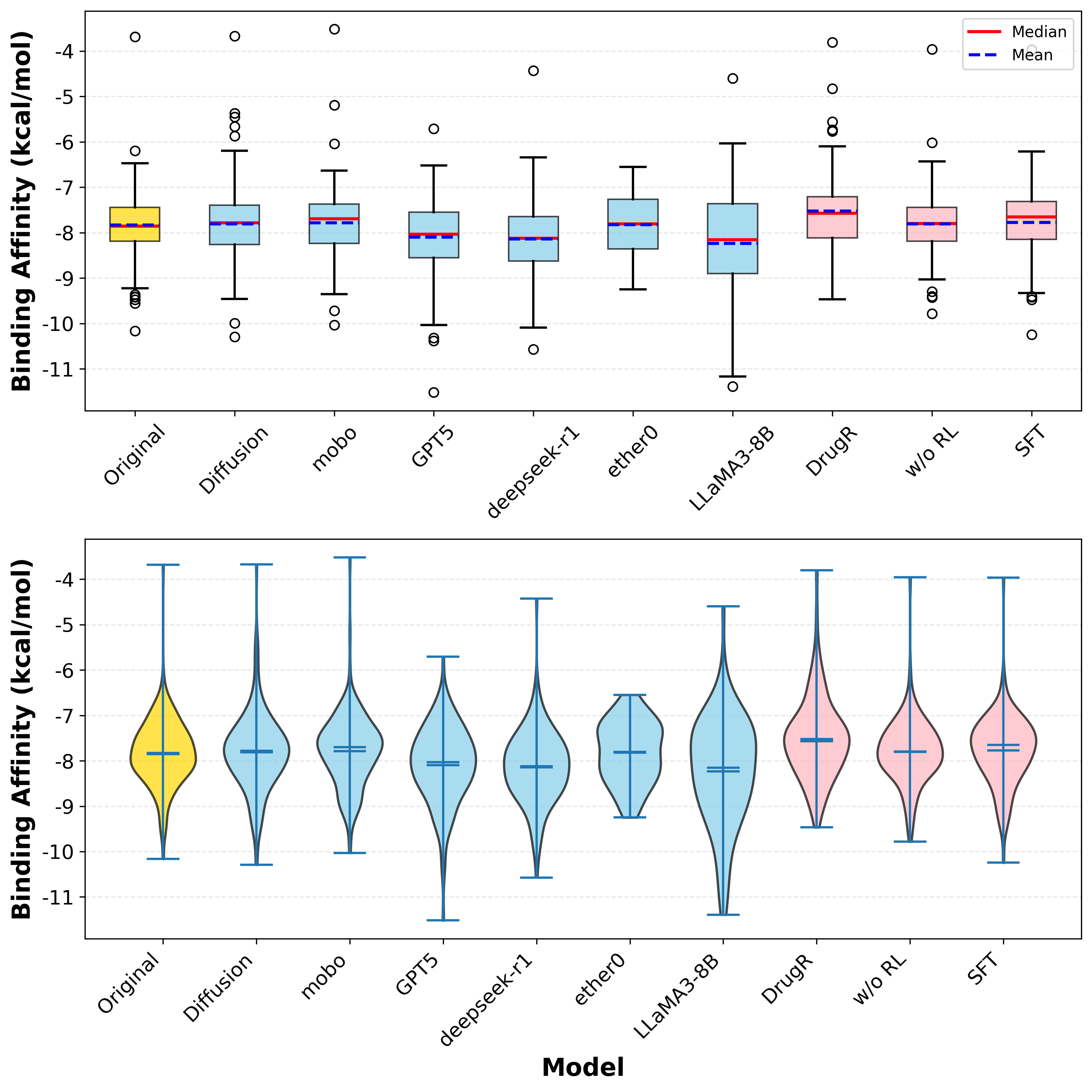}
    \caption{
    Binding energy comparison on the \emph{corresponding target receptor} across different molecular optimization methods. (A) Distribution of predicted binding free energy (docking score, kcal/mol) for optimized molecules obtained by each method on their designated targets. (B) Mean predicted binding free energy with standard error bars. Lower values indicate stronger predicted binding.
    }
    \label{fig:cox2_binding}
\end{figure}

To further assess whether pharmacological optimization compromises target-level functionality, we evaluate the binding affinity of optimized molecules against the corresponding targets of the given original molecules. Due to issues with the source of the data collection, some anti-inflammatory molecules lack precise target information; therefore, we only evaluate molecules with relevant target records. Figure~\ref{fig:cox2_binding} shows the distribution (A) and average (B) of the binding free energies predicted by different optimization methods. Notably, docking scores $\le$ -6~kcal/mol can be considered satisfactory under a commonly used empirical cutoff for strong and likely binding \cite{khadem2025cad,zhang2024deepgpcr}.

According to the results in Figure~\ref{fig:cox2_binding}, DrugR significantly improves ADMET-related properties while maintaining target binding affinity, generating approximately 95.84\% of ``satisfying design''. This ratio is roughly the same as the distribution of the original molecular distribution. Compared with the backbone model, DrugR generates more concentrated results after targeted training, with fewer extreme cases. As for other baselines, we have discussed the phenomenon of designing structures that deviate significantly from the given molecules and are closer to common drugs. The molecules generated by these models perform better in terms of binding affinity compared to the original molecules, which also supports this hypothesis.

Naturally, we also observe that DrugR generates a small subset of poor results in terms of bining affinity. By observing the model performance at different stages of RL, we find that optimizing certain drug properties (e.g., hepatotoxicity) inevitably involves the substitution of the active ingredient. Therefore, there is a trade-off between binding affinity and optimization score. We achieve a basic balance through multi-granular RL, but this also suggests that the current method may tend to limit its effectiveness in optimizing drug properties.

\subsection{Adaptation Capability}

\label{subsec:adaptation}

To evaluate the generalization ability of our approach under domain shift, we conduct an out-of-distribution (OOD) adaptation experiment across therapeutic areas. During the training stages, all models are exposed only to molecules from three drug categories as mentioned before. We then evaluate the models on a held-out anticancer dataset (200 pieces), which exhibits substantially different chemical scaffolds, pharmacological targets, and ADMET trade-offs. To adapt to this new domain, we fine-tune the models using only 50 anticancer training examples and report the post-adaptation performance on the held-out set.
Table~\ref{tab:ood} reports the OOD optimization performance. Intuitively, the untrained baselines (ExLLM, ether0, and LLaMA3-8B) still perform poorly on new test data, both in terms of overall optimization score and the fingerprint similarity.

While our method performs better, its data-driven training approach encounters difficulties when generalizing to molecules outside the fit range, and the help provided by a small sample size is not significant.

However, requiring only a very small amount of data and computational power (equivalent to 10 minutes of training on a single A100), DrugR can quickly master the basic optimization capabilities for new drug categories. Especially when compared to models using only the SFT strategy, it is evident that our method, through continuous pre-training and RL stages,successfully enables the model to grasp some general paradigms and disciplinary common sense regarding drug molecule optimization, providing a foundation for its extrapolation to broader application scenarios.

\begin{figure*}[!t]
  \centering
  \includegraphics[width=\textwidth,trim=16 100 16 16,clip]{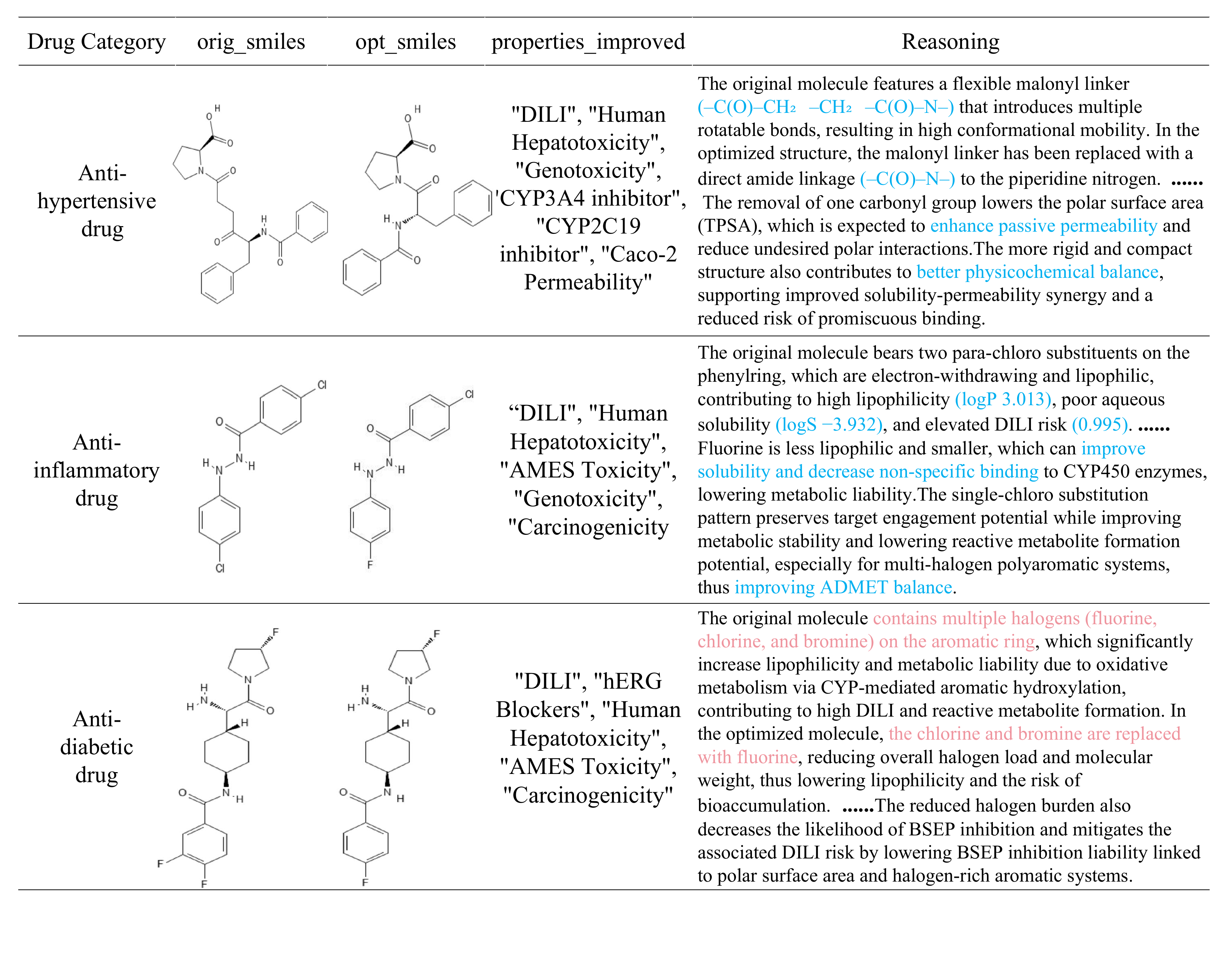}
  \vspace{-6pt}
  \caption{High-quality case studies for three therapeutic classes.}
  \label{fig:cases}
  \vspace{-10pt}
\end{figure*}

\subsection{Ablation and Case Analysis}
Table ~\ref{tab:main_results} also presents the results of the ablation experiments. Due to the time-consuming RL phase, we evaluate the effectiveness of the preliminary stages separately. 

The scores from the \textit{Pre-training}
show that the injected chemical knowledge may reduce general instruction following capability to some extent. However, comparing \textit{w/o RL} and pure \textit{SFT}, we can see that the former achieves a more comprehensive performance improvement with the presence of the pre-training. This indicates that domain-specific knowledge enhancement can lead to more stable inference and handle more flexible scenarios.
During the RL stage, we observe that the \textit{w/o RL} model scores significantly lower in both optimization and reasoning scores compared to the full setup. As mentioned before, in our automatically annotated data, the optimization score for the reference answer is only 0.165. This highlights the importance of reinforcement learning, which allows the model to freely explore and overcome the limitations of the training data distribution. Nevertheless, this exploration bias also led to a slight decrease in fingerprint similarity, meaning the model tended to generate molecules with more drastic variations.

We also conduct a \textit{w/o reasoning} setting, in which the model directly generates the optimized molecules without the explicit reasoning. 
The overall optimization score dropped sharply (to only 0.015). This is because, under the current training settings, it is difficult for LLMs to generate SMILES directly without a reasoning process. This indicates that explicit reasoning can help the model better utilize its world knowledge, bridging the gap between the model's inherent next token prediction approach and the specific demands of SMILES generation.

Fig.~\ref{fig:cases} presents representative case demonstrations across three therapeutic classes (anti-hypertensive, anti-inflammatory, and anti-diabetic), illustrating how the model connects structure edits between the original and optimized molecules to targeted ADMET/toxicity property improvements. In the rationale column, blue text indicates statements that are consistent with the depicted structural modifications and attribute property changes through plausible mechanistic mediators (e.g., altered flexibility or polarity affecting permeability and physicochemical balance). There also exist unsatisfying reasoning process as marked by red text, including structure-text mismatch (e.g., asserting substitutions absent from the structures) or mechanism skipping, where toxicity-related conclusions are claimed without specifying intermediate determinants such as ionization state and transporter affinity. This indicates possible directions for further research to enhance the grounding and consistency of explicit reasoning.

\section{Discussion}

Beyond predictive performance, an important motivation of our work is to make molecular optimization explainable by explicitly modeling the reasoning process that connects structural edits to downstream ADMET changes. DrugR is designed not only to propose improved molecules, but also to produce interpretable pharmacological rationales that can be inspected, audited, and distilled into reusable scientific heuristics. New decision basis that does not appear in training data emerges (e.g. ``reduced metabolic soft spot and altered lipophilicity profile can improve clearance and reduce bioaccumulation, addressing high BCF and DILI risk''). In this way, explicit reasoning bridges data-driven optimization and mechanism-oriented understanding, offering a practical route to summarize design experience and to generate hypotheses about structure–property relationships. 

Still, our method has several limitations. Similarity filtering and validity checks can reduce extreme excursions, but more subtle distribution shift may persist; under long-horizon reinforcement learning, the policy may exploit biases in the surrogate scoring function, producing high-scoring molecules that may not generalize outside the surrogate's validity domain~\cite{langevin2022failure}. Additionally, the optimization pipeline relies on a single learned evaluator (ADMETLab), whose predictions may depend on training data coverage and endpoint-specific reliability. Finally, the evaluation is performed primarily in silico to enable scalable comparisons, but it cannot rule out reward hacking~\cite{yoshizawa2025rewardhacking} and cannot guarantee that proxy-level gains translate to experimental improvements. These limitations suggest several directions for future work, including more fine-grained reward functions design, better evaluator towards larger molecular distribution region, and closed-loop optimization incorporating wet-lab experiments.

\section{Methods}
\label{sec:method}

\subsection{Related Work}

In its early stages, molecular optimization was commonly formulated as a black-box search problem, tackled with traditional machine learning and heuristic methods.
Conventional approaches such as Bayesian optimization (BO), genetic algorithms (GA), and GB-GA and its extensions are proven effective for single-property maximization tasks ~\cite{jensen2019gbga,nigam2019augmenting,tripp2021fresh}. Beyond these, methods based on Markov Chain Monte Carlo (MCMC) sampling and Monte Carlo tree search offered probabilistic exploration of chemical space ~\cite{xie2021mars,sun2022molsearch}.

To address practical scenarios involving multiple objectives, multi-objective optimization (MOO) techniques equipped with learned encoders or graph representations were developed to identify Pareto-optimal solutions ~\cite{liu2025mlps,verhellen2022graphpareto}.
Concurrently, reinforcement learning emerged as a distinct approach, with pipelines such as REINVENT and its variants (e.g., DyMol~\cite{ijcai2024p666}, Genetic-GFN~\cite{kim2024geneticgflownets}) being developed to train generative policies via reward signals~\cite{olivecrona2017molecular,jin2020multiobjective,guo2024augmented}. These methods often integrate techniques like experience replay or evolutionary operators to improve their efficacy on multi-objective tasks.

Although solid results were achieved on certain tasks~\cite{gao2022sampleefficiency}, these early methods largely relied on iterative function evaluations and hand-designed rewards, making it difficult to incorporate domain priors or chemist intuition. They also commonly required significant re-engineering when optimization objectives or constraints changed, highlighting limitations in flexibility and prior knowledge integration.

Deep generative models advanced the field by learning molecular distributions and producing higher-quality candidate proposals. Autoencoding and structured-latent approaches (e.g., JTVAE, VJTNN, and DST) capture regularities in scaffolds and substructures~\cite{pmlr-v80-jin18a,jin2019graph2graph,fu2022dst}, while diffusion and transformer-based models (e.g., MOOD, MolGPT, and MOLGEN) further improve sample fidelity and cross-domain applicability~\cite{pmlr-v202-lee23f,bagal2022molgpt,fang2024molgen}. Latent space optimization (LSO) suggests that multi-objective search in the learned latent space can be effective for deep generators~\cite{abeer2024molso}. However, these approaches also have limitations, such as heavy dependence on large-scale data and reward signals. In out-of-distribution settings, such as molecular optimization involving distinct structural features or novel property objectives, these systems can struggle to produce chemically valid outputs.
~\cite{gao2022sampleefficiency,liu2025mlps,verhellen2022graphpareto}.

With the development of pre-training technology, LLMs show impressive capabilities across various domains including biochemistry. 
Research has explored multiple paradigms for harnessing LLMs in molecular optimization.
One strategy employs pre-trained LLMs as optimizers, utilizing their inherent knowledge and reasoning to navigate chemical spaces efficiently without further training~\cite{ai4science2023impact,wu2024ecsurvey,brown2020gpt3}. 
Another approach involves fine-tuning domain-specialized LLMs to achieve superior performance in specific molecular understanding or generation tasks~\cite{zeng2022deep,zeng2024chatmol}.
Furthermore, LLMs can also be regarded as strong assistants for traditional tools and models. For instance, OPRO and LMEA treat LLMs as crossover/mutation operators in GA-style loops, balancing exploitation and exploration through prompt design and temperature control~\cite{yang2024opro,liu2024lmea}.Specific to molecular tasks, GA-style LLM-in-the-loop methods often use task-specific prompt templates to steer LLM-based mutations and crossovers, together with validity checks or filtering to improve the quality of generated molecules~\cite{wang2024llm_cmo}.
Meanwhile, several studies report that GA+LLM pipelines can be efficient and competitive with standalone LLMs and traditional multi-objective optimization (MOO) methods when coupled with well-structured workflows~\cite{liu2023ael,liu2024eoh,liu2024llm_moea,huang2024blackbox_llm,brahmachary2024leo}, indicating that LLMs are promising optimizers for numerical optimization, coding, and planning problems.

Complementary to treating LLMs as external optimizers or workflow components, a parallel line of related work focuses on chemistry-specialized and chemical-reasoning LLMs that directly improve molecular understanding, structured chemical knowledge usage, and multi-step scientific reasoning.
For example, ether0 post-trains a general LLM into a chemistry reasoning model that can follow natural-language rationales and output chemical structures, emphasizing data-efficient specialization via reinforcement learning on experimentally grounded chemistry tasks~\cite{narayanan2025ether0}. In contrast, ChemDFM builds a chemistry foundation model through large-scale domain corpus pre-training and instruction tuning to enhance chemistry dialogue, knowledge recall, and reasoning in diverse downstream tasks~\cite{zhao2024chemdfm}. ChemLLM further advances this direction by proposing a chemistry-dedicated LLM framework together with chemistry-specific instruction data and a benchmark suite, enabling systematic evaluation and improved performance across broad chemistry task categories~\cite{zhang2024chemllm}.
Despite their progress, existing chemistry-specialized LLMs often struggle to translate improved reasoning into reliable structure edits that consistently satisfy multi-objective constraints, particularly when supervision for property-changing modifications is scarce. In addition, many approaches emphasize task-specific understanding or generation but provide limited mechanisms for explicitly linking each molecular modification to the resulting ADMET trade-offs, which reduces interpretability and hinders robust generalization to new drug types or property objectives.

\subsection{Evaluation Details}\label{section:reward}

\begin{figure*}[!t]
  \centering
  \includegraphics[width=\textwidth]{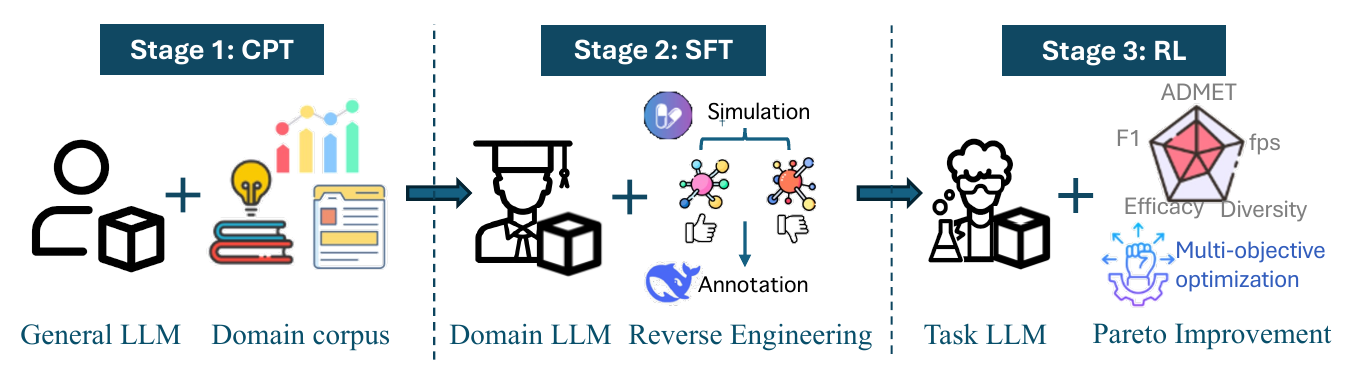}
  \caption{The three-stage training process of DrugR.}
  \label{fig:framework}
\end{figure*}

\textbf{Overall Optimization Score}. ADMET (Absorption, Distribution, Metabolism, Excretion, and Toxicity) is commonly used to formalize this objective by characterizing the developability and safety profile of drug candidates~\cite{zhang2025computational_toxicology_admet}. A common design in ADMET-oriented scoring is to compress multiple endpoint predictions into a single scalar, which effectively recasts molecular optimization as maximizing an absolute objective that is assumed to be smooth and largely monotonic. ADMET-score~\cite{guan2019admet} exemplifies this paradigm by aggregating 18 admetSAR-predicted endpoints into a composite drug-likeness index through a weighted combination of binarized beneficial versus harmful outcomes, with weights derived from endpoint importance, predictive accuracy, and a usefulness index. Such absolute endpoint scores are convenient for ranking candidates, yet they only weakly reflect the semantics of iterative optimization.

We compute the main optimization score by emphasizing improvements on the original ADMET liabilities and penalizing newly introduced liabilities. 
Given an original--optimized pair $(x_o, x_p)$, we obtain endpoint predictions and derive the liability set $\mathcal{B}(x)$ using the criteria in Tables~\ref{tab:admet-criteria-main}--\ref{tab:admet-monotonic-no-threshold}. 
Let $\Delta_f(o_f,p_f)$ denote the direction-aware improvement for endpoint $f$ (range-target, threshold-based, or monotonic), and clip each endpoint's contribution to $[-M, M]$ with $M=2.0$. 
We aggregate improvements as
\begin{equation}
\begin{aligned}
R_{\mathrm{main}}(x_o,x_p)
=\mathrm{Norm}\!\Bigg(
&\sum_{f\in \mathcal{C}} a_f(x_o)\,
\mathrm{clip}\!\big(\Delta_f(o_f,p_f),-M,M\big) \\
&\;-\mu\,\big|\mathcal{B}(x_p)\setminus \mathcal{B}(x_o)\big|
\Bigg).
\end{aligned}
\end{equation}
where $\mathcal{F}_o$ and $\mathcal{F}_p$ are the predicted endpoint sets for $x_o$ and $x_p$, and $\mathcal{C}=\mathcal{F}_o\cap\mathcal{F}_p$.
We use a fixed gating weight $a_f(x_o)=1$ if $f\in\mathcal{B}(x_o)$ (an original liability), and $a_f(x_o)=\tfrac{1}{2}$ otherwise, so that improvements on non-liability endpoints contribute at half weight. We set $\mu=0.3$ as a fixed penalty per newly introduced liability. For threshold-based endpoints, we add a constant bonus $\beta$ when the prediction crosses the threshold to the desirable side; for range-target endpoints, we add a constant bonus $\gamma$ when the prediction enters the target interval. Throughout, we set $\beta=2.5$ and $\gamma=3.5$ (Table~\ref{tab:admet-criteria-main}); monotonic endpoints instead use a continuous relative-improvement term without discrete bonuses.

\begin{table}[t]
  \centering
  \caption{Target criteria for key ADMET indicators used in the reward function.}
  \label{tab:admet-criteria-main}
  \small
  \setlength{\tabcolsep}{4pt}
  \renewcommand{\arraystretch}{1.08}
  \begin{tabularx}{\columnwidth}{@{} >{\raggedright\arraybackslash}X r c c @{}}
    \toprule
    \textbf{Indicator} & \textbf{Target/Thr.} & \textbf{Dir.} & \textbf{Bonus} \\
    \midrule
    \multicolumn{4}{@{}l}{\textit{Range targets}} \\[-2pt]
    logP & $[1.0,\,3.0]$ & $\leftrightarrow$ & $+\gamma$ \\
    TPSA & $[20.0,\,130.0]$ & $\leftrightarrow$ & $+\gamma$ \\
    MW & $[150.0,\,500.0]$ & $\leftrightarrow$ & $+\gamma$ \\
    \midrule
    \multicolumn{4}{@{}l}{\textit{Threshold-based}} \\[-2pt]
    Caco-2 permeability & $-5.15$ & $\downarrow$ & $+\beta$ \\
    F50\% & $0.50$ & $\downarrow$ & $+\beta$ \\
    CYP3A4 inhibitor & $0.50$ & $\downarrow$ & $+\beta$ \\
    CYP2D6 inhibitor & $0.50$ & $\downarrow$ & $+\beta$ \\
    P-gp substrate & $0.50$ & $\downarrow$ & $+\beta$ \\
    hERG blockers & $0.80$ & $\downarrow$ & $+\beta$ \\
    DILI & $0.80$ & $\downarrow$ & $+\beta$ \\
    Human hepatotoxicity & $0.80$ & $\downarrow$ & $+\beta$ \\
    AMES toxicity & $0.80$ & $\downarrow$ & $+\beta$ \\
    Genotoxicity & $0.80$ & $\downarrow$ & $+\beta$ \\
    Drug-induced neurotoxicity & $0.80$ & $\downarrow$ & $+\beta$ \\
    QED & $0.34$ & $\downarrow$ & $+\beta$ \\
    SA score & $0.50$ & $\downarrow$ & $+\beta$ \\
    GASA & $0.50$ & $\downarrow$ & $+\beta$ \\
    Lipinski rule & $0.50$ & $\downarrow$ & $+\beta$ \\
    \bottomrule
  \end{tabularx}

  \vspace{2pt}
  {\footnotesize \textbf{Dir.} $\downarrow$: lower is better; \ $\leftrightarrow$: best within the target range.
  \ \textbf{Bonus}: $+\beta$ for crossing the threshold to the desirable side, $+\gamma$ for entering the target range
  (we set $\beta=2.5$ and $\gamma=3.5$).}
\end{table}

\begin{table}[t]
  \centering
  \caption{Monotonic ADMET indicators without explicit thresholds used in the reward function.}
  \label{tab:admet-monotonic-no-threshold}
  \small
  \setlength{\tabcolsep}{6pt}
  \renewcommand{\arraystretch}{1.10}
  \begin{tabularx}{\columnwidth}{@{} >{\raggedright\arraybackslash}X c c @{}}
    \toprule
    \textbf{Indicator} & \textbf{Dir.} & \textbf{Reward} \\
    \midrule
    HLM stability & $\uparrow$ & Relative \\
    logS & $\uparrow$ & Relative \\
    logD7.4 & $\uparrow$ & Relative \\
    Flexibility & $\uparrow$ & Relative \\
    Fsp3 & $\uparrow$ & Relative \\
    \bottomrule
  \end{tabularx}

  \vspace{2pt}
  {\footnotesize \textbf{Dir.} $\uparrow$: higher is better.
  \ \textbf{Reward}: continuous relative-improvement term (no discrete threshold/range bonus).}
\end{table}

\textbf{Binding Affinity}. A standardized docking pipeline is used to compute this indicator~\cite{meng2011docking}. Ligand 3D conformers are generated from SMILES with RDKit (ETKDG)~\cite{riniker2015etkdg} and minimized using MMFF94~\cite{halgren1996mmff}. The resulting structures are converted to PDBQT and assigned Gasteiger charges with Open Babel~\cite{oboyle2011openbabel,gasteiger1980charges}, followed by docking with AutoDock Vina under a fixed search space and consistent protocol~\cite{trott2010vina}. The Vina score (kcal/mol) is used for relative comparison across candidates, with lower values indicating stronger predicted binding.

Targets are selected based on drug category and established mechanisms of action (Table~\ref{tab:target_sets}), and the identical workflow is applied per target to ensure comparability. During evaluation, improvement in the overall optimization objective is treated as primary, while docking scores and molecular fingerprint similarity are constrained to remain within predefined ranges (see \S4.5), reflecting lead optimization settings that seek global pharmacological gains without compromising target engagement or disrupting the core scaffold.


\begin{table*}[t]
\centering
\small
\setlength{\tabcolsep}{5pt}
\caption{Category-specific target sets used for binding affinity evaluation.}
\label{tab:target_sets}
\begin{tabular}{|
>{\centering\arraybackslash}m{0.26\textwidth}|
>{\centering\arraybackslash}m{0.18\textwidth}|
m{0.52\textwidth}|}
\hline
\textbf{Drug category} & \textbf{Target set} & \textbf{Representative drugs (examples)} \\
\hline
Anti-inflammatory (NSAIDs) & COX1, COX2 &
aspirin, ibuprofen, diclofenac, celecoxib \\
\hline
Antihypertensive (ACEi/ARB/$\beta$-blockers) & ACE, AGTR1, ADRB1, ADRB2 &
captopril, losartan, propranolol, metoprolol \\
\hline
Antidiabetic (DPP4i/GLP1R agonists) & DPP4, GLP1R &
sitagliptin, linagliptin, exenatide, semaglutide \\
\hline
\end{tabular}
\end{table*}

\textbf{LMS and richness score}. The judger outputs normalized sub-scores that are combined through weighted aggregation with explicit penalties for critical failures such as edit–rationale contradictions or severe chemical misconceptions; empty or uninformative rationales receive zero. This construction yields a bounded and stable metric suitable for reinforcement-learning settings in which previously unseen reasoning patterns may arise.

For the reasoning-richness score, rather than rewarding proximity to templated explanations or tolerating highly off-distribution claims, this signal measures the semantic distance between a generated rationale and a reference space constructed from validation rationales via sentence-embedding prototypes, and converts this distance into a bounded $[0,1]$ score using a unimodal mapping that peaks at an empirically chosen intermediate deviation while penalizing both over-proximity and excessive drift.
The prompt templates used to elicit and standardize rationales for constructing the reference space and computing this score are provided in Supplementary Information, Section S1.

\subsection{Continual pre-training}
\label{sec:acke}

The Continual pre-training stage(CPT) equips the model with fine-grained chemical knowledge that can be verbalized as coherent rationales during downstream molecular optimization. Starting from LLaMA-3-8B-Instruct, domain adaptation is carried out via continued pretraining on a chemistry-centered mixture, while general-domain corpora are retained to preserve broad linguistic competence and maintain instruction-following behavior.

The training mixture integrates four data sources with complementary inductive signals. An internal ChemicalQA collection (\(\sim\)150K) provides conceptual supervision over functional groups, structural motifs, structure interpretation, and SMILES-related queries~\citep{chemqa_internal}. In parallel, MoleculeNet benchmarks (\(\sim\)160K) deliver property- and activity-driven supervision for ADMET/toxicity and bioactivity prediction~\citep{wu2018moleculenet}. General conversational alignment is supported by UltraChat-200K (\(\sim\)200K)~\citep{ding-etal-2023-enhancing}, and a large CPT text corpus (\(\sim\)300K) stabilizes language modeling under long sequences~\citep{cpt_internal}. According to corpus statistics, the MoleculeNet subset is English-only (132.5 tokens/example on average), whereas UltraChat and CPT average 521.2 and 617.9 tokens with negligible CJK proportions (0.07\% and 0.14\%, respectively).

\subsection{Supervised fine-tuning}
\label{sec:crl}

\subsubsection{Explicit Reasoning Dataset}

We present an explicit-reasoning dataset for molecular optimization, in which each example includes a rationale describing structural edits and their effects on ADMET properties. The dataset contains 4,855 samples, each pairing an input molecule (with its ADMET profile) with a detailed reasoning trace followed by an optimized SMILES string.

To generate these reasoning annotations at scale, we build a closed-loop reverse-engineering pipeline. Instead of manually labeling optimization rationales, the pipeline starts from established drug molecules, produces structurally comparable candidates, verifies whether optimization occurs via ADMET evaluation, and then works backward from outcomes to explanations. This process yields structured, explanation-centric supervision for molecular optimization.

Starting molecules are curated from the ChEMBL database\footnote{\url{https://www.ebi.ac.uk/chembl/}},
focusing on bioactive compounds with clear therapeutic targets across three representative drug categories: anti-inflammatory drugs (e.g., targeting COX-1/COX-2), antihypertensive drugs (e.g., targeting ACE and related cardiovascular proteins), and antidiabetic drugs (e.g., targeting diabetes-relevant proteins). As a result of real-world drug discovery practices, the resulting dataset is inherently imbalanced across categories, reflecting natural differences in data availability and chemical diversity. All molecules are standardized and validated to ensure chemical consistency.

Starting from a given molecule, DeepSeek-R1 is employed to propose a set of structurally related candidates, leveraging its strong multi-step reasoning capability and long-horizon generation. Fingerprint-based similarity filtering is applied to enforce structural comparability, ensuring that subsequent evaluations reflect meaningful local edits rather than chemically unrelated alterations. For each retained molecular pair, ADMET properties are predicted with ADMETLab, covering absorption, metabolism, toxicity, as well as synthetic accessibility and drug-likeness.

To determine the optimization direction, we adopt a unified threshold-based rule. A candidate is retained if and only if it exhibits a sufficiently large improvement on the originally disadvantaged ADMET property relative to its source molecule, while preserving structural consistency with a fingerprint similarity greater than 0.6.

For molecular pairs that satisfy the improvement criteria, the reverse-engineering pipeline derives reasoning supervision by working backward from verified ADMET outcomes. Concretely, we compute a compact summary of property deltas between the source and candidate molecules and feed it, together with the two SMILES strings, into a chemistry-adapted language model to generate concise, mechanistically grounded rationales that connect structural edits to the observed ADMET changes. The prompting template used for rationale generation is provided in Supplementary Information S2.

Using this pipeline, we construct a molecular optimization reasoning dataset containing 4,855 samples. Overall, the reverse-engineering pipeline framework ensures that the resulting supervision is objective, interpretable, chemically consistent, and grounded in realistic drug optimization scenarios.

\begin{table}[t]
\centering
\small
\caption{Molecular complexity and diversity statistics.}
\label{tab:mol_stats_compact}
\begin{tabular}{lcc}
\toprule
\textbf{Metric} & \textbf{Mean / Value} & \textbf{Range / Definition} \\
\midrule
Heavy atoms & 24.12 & 2--43 \\
Rings & 3.70 & 0--8 \\
Rot. bonds & 3.13 & 0--20 \\
TPSA & 60.43 & 0.00--198.12 \\
cLogP & 3.48 & -2.28--9.63 \\
Unique SMILES & 3{,}863 / 4{,}826 & 80.05\% \\
Unique scaffolds & 1{,}117 / 4{,}826 & 23.15\% (Bemis--Murcko) \\
Mean Tanimoto & 0.1291 & 1{,}000 sampled molecules \\
Diversity index & 0.8709 & $1-\mathrm{mean\_sim}$ \\
\bottomrule
\end{tabular}
\end{table}

\subsubsection{Dataset Statistics}

We conduct a systematic statistical characterization of the constructed dataset. It contains 4,855 samples and is split into train/dev/test subsets with an 85\%/10\%/5\% ratio (4,126/485/244), enabling stable model selection and reliable generalization evaluation. Token-level statistics (Table~\ref{tab:mol_stats_compact}) indicate that most instances fit within common context windows, while a
non-trivial tail near the upper limit provides a useful stress test for long-context robustness.

From the molecular perspective, the dataset exhibits broad structural complexity
and chemical diversity, with wide coverage in atom counts, ring systems,
flexibility, and physicochemical properties (Table~\ref{tab:mol_stats_compact}). The
average pairwise similarity is low, indicating extensive coverage of chemical
space. The dataset spans three therapeutic categories (anti-inflammatory,
antihypertensive, and antidiabetic) and shows a markedly imbalanced distribution,
reflecting real-world data availability in curated bioactivity repositories.

Each sample is accompanied by a comprehensive ADMET profile consisting of 23
features spanning absorption, distribution, metabolism, excretion, and toxicity.
Threshold definitions and evaluation criteria for these properties are summarized
in Table~\ref{tab:admet-criteria-main}. Overall, the dataset combines long-context,
explanation-centric supervision with diverse molecular structures and rich ADMET
signals, making it well suited for studying molecule-centric instruction learning
and multi-objective optimization under realistic constraints.
Prompt templates and data generation/format specifications are provided in Supplementary Information S3.

\subsubsection{Implementation Details}
\label{sec:crl}

Supervised fine-tuning (SFT): We fine-tune the model on task data constructed via reverse data engineering to strengthen its capability to (i) recognize ADMET liabilities of a given molecule and (ii) design targeted structural modifications that address these weaknesses. Importantly, SFT is not limited to optimizing the final SMILES string; it also performs chemical reasoning learning, training the model to generate faithful, task-relevant rationales that explicitly link concrete edits (e.g., substituent replacement, polarity/lipophilicity adjustment, steric or hydrogen-bonding changes) to predicted property-level outcomes and potential trade-offs across objectives. This encourages the model to ground its decisions in mechanistic and property-aware arguments rather than relying on shallow template matching.

To balance SMILES validity and reasoning fidelity, we adopt a two-stage curriculum. Stage 1 supervises only SMILES generation, focusing on molecular syntax, canonicalization patterns, and stable structure representation, thus reducing invalid outputs and freeing capacity for higher-level reasoning later. Stage 2 resumes from the Stage-1 checkpoint and jointly supervises the generation of both rationales and optimized SMILES strings, aligning the model’s explanations with the causal structure of molecular optimization and improving consistency between the described edits and the predicted ADMET changes.

\subsection{Self-balanced Multi-granular Reinforcement Learning}
\label{sec:self_balanced_rl}

In our molecular optimization setting, a single response contains two semantically distinct segments: a Reasoning segment that diagnoses ADMET liabilities and motivates concrete chemical edits, and an Optimized SMILES segment that instantiates these edits as a chemically valid molecule. Both segments are governed by multi-objective criteria. The Reasoning segment is optimized with three objectives capturing diagnostic correctness and scientific coherence (target-property F1, reasoning lms score, and reasoning richness), whereas the SMILES segment is optimized with three objectives capturing optimization progress under feasibility constraints (overall optimization score, fingerprint similarity, and binding energy). Because these objectives differ in reward scale and convergence rate, naïvely aggregating them tends to induce objective domination and starvation, in which faster-improving channels overwhelm slower yet crucial ones.

\begin{figure}[!t]
  \centering
  \includegraphics[width=\linewidth]{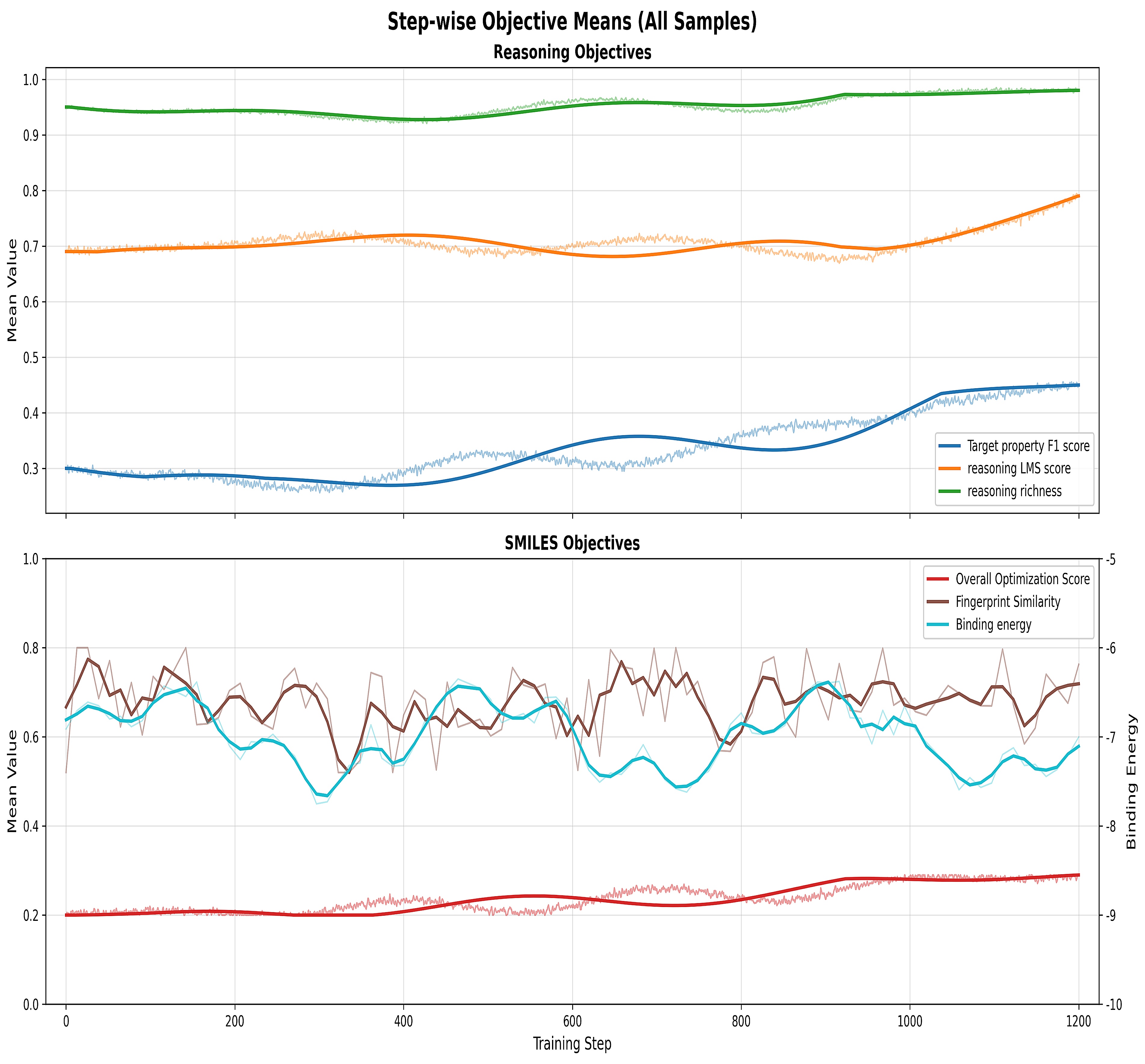}
  \caption{\textbf{Multi-objective training dynamics under Pareto-aware self-balancing.}
  Step-wise objective means over training iterations (computed over all sampled trajectories).
  \emph{Top:} reasoning objectives (Target-property F1, LMS, and reasoning richness).
  \emph{Bottom:} SMILES-side objectives (overall optimization score, fingerprint similarity, and binding energy; lower binding energy is better and is shown on the right axis).
  The curves demonstrate stable, non-degenerate co-improvement across all six objectives, consistent with our two-level balancing scheme.}
  \label{fig:pareto}
\end{figure}

\begin{table}[t]
\centering
\small
\setlength{\tabcolsep}{7pt}
\begin{tabular}{rcccc}
\toprule
Step & $p50(w)$ & $p90(w)$ & $p99(w)$ & Frontier ratio \\
\midrule
0    & 0.990 & 1.160 & 1.270 & 0.160 \\
300  & 0.975 & 1.179 & 1.276 & 0.207 \\
600  & 0.967 & 1.164 & 1.301 & 0.203 \\
900  & 0.959 & 1.178 & 1.269 & 0.250 \\
1200 & 0.966 & 1.135 & 1.228 & 0.297 \\
\bottomrule
\end{tabular}
\caption{Pareto sample-weight dynamics (mean-normalized within each batch). Quantiles summarize the tail behavior of $w_i$ and the frontier ratio reports $|\mathcal{P}|/N$.}
\label{tab:pareto_weight}
\end{table}

To address this issue, we adopt a two-level adaptive self-balancing mechanism that redistributes learning signal across trajectories while stabilizing inter-objective training dynamics. At the sample level, we perform Pareto-aware reweighting within each on-policy rollout batch. Concretely, we construct per-trajectory objective vectors and estimate an empirical Pareto set $\mathcal{P}$ via dominance relations. Trajectories on $\mathcal{P}$ receive a multiplicative boost, whereas dominated trajectories are softly adjusted based on their minimum distance to Pareto anchors through an exponential decay. This produces a sample-wise scalar weight that uniformly scales all reward channels (including token-level rewards) for that trajectory. The uniform scaling is critical: it increases the gradient contribution of multi-objective-consistent rollouts without distorting the within-trajectory proportions among reward channels. Formally, for a batch of $N$ trajectories, we define a raw weight $\tilde w_i$ and its mean-normalized counterpart $w_i$ as
\begin{equation}
\tilde w_i=
\begin{cases}
\beta, & i\in\mathcal{P},\\[2pt]
\displaystyle\prod_{m=1}^{M}\Bigl(1+(\beta-1)\exp(-\lambda\, d_i^{(m)})\Bigr), & i\notin\mathcal{P},
\end{cases}
\qquad
w_i=\frac{N\,\tilde w_i}{\sum_{j=1}^{N}\tilde w_j},
\label{eq:pareto_weight}
\end{equation}
where $\beta$ is the Pareto-front boost factor, $\lambda$ controls the distance decay, and $d_i^{(m)}$ denotes the minimum distance from trajectory $i$ to the Pareto anchors under the $m$-th distance metric. In our implementation, $M=2$ to reflect two complementary distance corrections computed in different subspaces. The normalization ensures $\frac{1}{N}\sum_{i=1}^{N} w_i = 1$, so the procedure reallocates importance across trajectories rather than changing the overall update magnitude.

While sample-level Pareto reweighting improves the quality of training signal, it does not fully eliminate imbalance caused by heterogeneous learning dynamics across objectives. Therefore, we further introduce batch-level adaptations driven by batch statistics. We compute group-level scaling factors for the Reasoning group and the SMILES group from aggregated batch performance, increasing the relative emphasis on the group that lags behind. In parallel, we apply a capped, channel-wise shortfall boost to objectives whose batch means fall below predefined targets, with a smooth exponent controlling aggressiveness and an explicit cap preventing over-correction. This second level mitigates domination effects arising from mismatched convergence rates and stabilizes training in the presence of heterogeneous rewards, thereby reducing sensitivity to per-channel manual calibration.

We evaluate effectiveness at both outcome and mechanism levels. In terms of outcomes, Fig.~\ref{fig:pareto} shows that the steps of all six objectives remain well-behaved without degenerate collapse, while showing sustained progress in training. In particular, the reasoning-side F1 score, which typically constitutes a performance bottleneck, exhibits a consistent improvement in later training stages. In parallel, the SMILES-side overall optimization score increases steadily, without inducing systematic degradation in fingerprint similarity or violations of binding-related constraints. Mechanism-wise, we monitor the distribution of Pareto sample weights throughout training. Since sample weights are mean-normalized within each batch, shifts in their quantiles (e.g., $p50$, $p90$, $p99$) directly reflect how strongly training emphasizes frontier-consistent trajectories, and the Pareto-set coverage ratio tracks the prevalence of multi-objective-consistent rollouts. Table~\ref{tab:pareto_weight} reports representative quantiles and frontier ratios, confirming that the reweighting procedure is active and evolves over the course of training. Collectively, these results support that the proposed self-balancing strategy yields stable, coordinated improvements across objectives while reducing reliance on extensive per-channel manual weight tuning.


\subsection{Model and Baselines}

To comprehensively evaluate the proposed framework, we compare against a diverse set of baselines
covering (i) classical non-LLM optimization methods, (ii) general-purpose LLMs, (iii) chemistry-adapted
molecular reasoning/generation models, and (iv) ablations of our framework. This taxonomy mirrors the
groups shown in Table~\ref{tab:main_results} and is intended to isolate the contributions of explicit
reasoning, reinforcement learning, and multi-objective reward design.

\paragraph*{Traditional (non-LLM) optimization methods.}
We first include two representative classical baselines that optimize molecular properties without
language reasoning:
MOBO, a Bayesian optimization pipeline, and Diffusion, a generative diffusion-based optimizer with 190M parameters. These methods serve as strong non-LLM references for optimization quality and structure
preservation under identical evaluation metrics.
\paragraph*{General-purpose LLM baselines.}
We evaluate general-purpose LLMs that are not specifically tailored to molecular optimization under
our objective, including GPT5 and DeepSeek-R1. These baselines reflect the capability
of off-the-shelf LLMs (with or without generic RL training) when directly prompted to produce
reasoned molecular edits.

\paragraph*{Chemistry-adapted reasoning/generation models.}
We further compare with LLM-based molecular models that are explicitly designed or adapted for
chemistry tasks, including ChemDFM-13B, ExLLM, Ether0-24B, and LLaMA3-8B.
These models represent chemistry-specialized architectures or domain-adapted LLMs, and provide a
strong reference point for molecular reasoning and generation beyond general-purpose LLMs.

\paragraph*{Backbone models for DrugR.}
For our method, we instantiate DrugR with several open-source backbones, including
Qwen3-8B, Mistral-7B, LLaMA3-8B, and LLaMA3.2-3B. These backbones
span 3B--8B parameters and vary in architectural depth, enabling analysis of how model capacity
affects reasoning quality and optimization performance. All variants follow autoregressive decoding
and generate the reasoning trace and optimized SMILES in a single sequence.

\paragraph*{Ablations of DrugR.}
To isolate the contributions of key components, we include the following ablations (Table~\ref{tab:main_results}):
(i) w/o RL, which removes GRPO and trains only with supervised objectives;
(ii) Pre-training, which uses domain pre-training without task-level alignment;
(iii) SFT, which fine-tunes on the explicit reasoning data via maximum likelihood;
and (iv) w/o reasoning, which disables reasoning generation and directly outputs optimized
SMILES.
Together, these ablations quantify the impact of explicit reasoning supervision and reinforcement
learning under the same multi-objective evaluation protocol.

\subsection{Training Settings}

We adopt GRPO as the reinforcement learning algorithm. Pareto-based dynamic reweighting is applied every 100 batches. All reward components
are normalized to compatible ranges to avoid scale dominance. We use AdamW for optimization, with a learning rate of $1\times10^{-5}$ for reinforcement learning. Each RL batch contains 16-32 rollouts depending on the model size.

For CPT, we use a learning rate of $5\times10^{-5}$.
Training is performed with a micro-batch size of $1$ and gradient accumulation, resulting in an
A100-equivalent effective batch size of $128$ sequences per update.
We set the maximum sequence length to $3072$ and enable sequence packing to improve token utilization.
Continual pre-training is run for a total of $2600$ steps.

We adopt a two-stage mixed-chemistry SFT schedule.
Starting from the pre-trained checkpoint at step $1200$, we first run a SMILES-only stage for
$800$ steps, followed by a reasoning+SMILES stage for an additional $1200$ steps.
Throughout this SFT process, we use a learning rate of $2\times10^{-5}$.
Training uses a micro-batch size of $2$ and gradient accumulation, resulting in an
A100-equivalent effective batch size of $64$ sequences per update (without revealing the underlying
hardware parallelism).

\section*{Conclusion}
This paper presents an explicit-reasoning framework for molecular drug optimization that targets ADMET improvement under a strict structural-similarity constraint. The approach integrates chemistry-oriented continued pretraining, a closed-loop reverse-engineering pipeline that converts verified property gains into explanation-centric supervision, and reinforcement learning for reasoning-guided candidate refinement. Empirically, the resulting system improves disadvantaged properties while maintaining fingerprint similarity above 0.6, and produces rationales that link concrete structural edits to predicted ADMET changes. Remaining limitations primarily stem from reliance on surrogate property predictors; future work will incorporate stronger evaluators and experimental validation, and extend coverage to broader therapeutic scopes.

\section*{Author contributions}
Haoran Liu: Conceptualization, Methodology, Investigation, Data curation, Formal analysis, Visualization.
Zheni Zeng: Supervision, Project administration, Funding acquisition, Conceptualization, Methodology \& editing.Yukun Yan and Yuxuan Chen:Investigation, Data curation, Software, Validation, Resources, Visualization, Writing.

\section*{Conflicts of interest}
There are no conflicts to declare.

\section*{Data availability}
The code and data used in the study are publicly available from the GitHub repository:
https://github.com/Haoranliu-lab/DrugR-main

\balance

\bibliography{rsc} 

@article{grattafiori2024llama3,
  title   = {The LLaMA 3 Herd of Models},
  author  = {Grattafiori, Aaron and Dubey, Abhimanyu and Jauhri, Abhinav and others},
  journal = {arXiv preprint arXiv:2407.21783},
  year    = {2024},
  url     = {https://arxiv.org/abs/2407.21783}
}

@article{rogers2010ecfp,
  title   = {Extended-connectivity fingerprints},
  author  = {Rogers, David and Hahn, Mathew},
  journal = {Journal of Chemical Information and Modeling},
  volume  = {50},
  number  = {5},
  pages   = {742--754},
  year    = {2010}
}

@article{bajusz2015tanimoto,
  title   = {Why is Tanimoto index an appropriate choice for fingerprint-based similarity calculations?},
  author  = {Bajusz, D{\'a}vid and R{\'a}cz, Anita and H{\'e}berger, K{\'a}roly},
  journal = {Journal of Cheminformatics},
  volume  = {7},
  number  = {1},
  pages   = {20},
  year    = {2015}
}

@article{admetlab3,
  title   = {ADMETlab 3.0: an updated comprehensive online ADMET prediction platform enhanced with broader coverage, improved performance, API functionality and decision support},
  author  = {Fu, Li and Shi, Shaohua and Yi, Jiacai and others},
  journal = {Nucleic Acids Research},
  year    = {2024},
  volume  = {52},
  number  = {W1},
  pages   = {W422--W431},
  doi     = {10.1093/nar/gkae236}
}

@article{guo2025deepseek,
  title={Deepseek-r1: Incentivizing reasoning capability in llms via reinforcement learning},
  author={Guo, Daya and Yang, Dejian and Zhang, Haowei and Song, Junxiao and Zhang, Ruoyu and Xu, Runxin and Zhu, Qihao and Ma, Shirong and Wang, Peiyi and Bi, Xiao and others},
  journal={arXiv preprint arXiv:2501.12948},
  year={2025}
}

@article{jensen2019gbga,
  title   = {A graph-based genetic algorithm and generative model/Monte Carlo tree search for the exploration of chemical space},
  author  = {Jensen, Jan H.},
  journal = {Chemical Science},
  year    = {2019},
  volume  = {10},
  number  = {12},
  pages   = {3567--3572},
  doi     = {10.1039/C8SC05372C},
  publisher = {The Royal Society of Chemistry}
}

@misc{nigam2019augmenting,
  title         = {Augmenting Genetic Algorithms with Deep Neural Networks for Exploring the Chemical Space},
  author        = {Nigam, AkshatKumar and Friederich, Pascal and Krenn, Mario and Aspuru-Guzik, Al{\'a}n},
  year          = {2019},
  eprint        = {1909.11655},
  archivePrefix = {arXiv},
  primaryClass  = {cs.LG},
  doi           = {10.48550/arXiv.1909.11655},
  url           = {https://arxiv.org/abs/1909.11655}
}

@inproceedings{tripp2021fresh,
  title     = {A Fresh Look at De Novo Molecular Design Benchmarks},
  author    = {Tripp, Austin and Simm, Gregor N. C. and Hern{\'a}ndez-Lobato, Jos{\'e} Miguel},
  booktitle = {NeurIPS 2021 AI for Science Workshop},
  year      = {2021},
  url       = {https://openreview.net/forum?id=gS3XMun4cl_}
}

@inproceedings{liu2025mlps,
  title     = {Multi-Objective Molecular Design Through Learning Latent Pareto Set},
  author    = {Liu, Yiping and Yang, Jiahao and Ren, Xuanbai and Zhang, Xinyi and Liu, Yuansheng and Song, Bosheng and Zeng, Xiangxiang and Ishibuchi, Hisao},
  booktitle = {Proceedings of the AAAI Conference on Artificial Intelligence},
  year      = {2025},
  volume    = {39},
  number    = {18},
  pages     = {19006--19014},
  doi       = {10.1609/aaai.v39i18.34092},
  url       = {https://ojs.aaai.org/index.php/AAAI/article/view/34092}
}

@article{verhellen2022graphpareto,
  title   = {Graph-based molecular Pareto optimisation},
  author  = {Verhellen, Jonas},
  journal = {Chemical Science},
  year    = {2022},
  volume  = {13},
  number  = {25},
  pages   = {7526--7535},
  doi     = {10.1039/D2SC00821A},
  publisher = {The Royal Society of Chemistry},
  url     = {https://pubs.rsc.org/en/content/articlelanding/2022/sc/d2sc00821a}
}

@inproceedings{xie2021mars,
  title     = {MARS: Markov Molecular Sampling for Multi-objective Drug Discovery},
  author    = {Xie, Yutong and Shi, Chence and Zhou, Hao and Yang, Yuwei and Zhang, Weinan and Yu, Yong and Li, Lei},
  booktitle = {International Conference on Learning Representations (ICLR)},
  year      = {2021},
  eprint    = {2103.10432},
  archivePrefix = {arXiv},
  primaryClass  = {q-bio.BM},
  doi       = {10.48550/arXiv.2103.10432},
  url       = {https://arxiv.org/abs/2103.10432}
}

@inproceedings{sun2022molsearch,
  title     = {MolSearch: Search-based Multi-objective Molecular Generation and Property Optimization},
  author    = {Sun, Mengying and Wang, Huijun and Xing, Jing and Chen, Bin and Meng, Han and Zhou, Jiayu},
  booktitle = {Proceedings of the 28th ACM SIGKDD Conference on Knowledge Discovery and Data Mining (KDD '22)},
  year      = {2022},
  pages     = {4724--4732},
  doi       = {10.1145/3534678.3542676},
  publisher = {Association for Computing Machinery},
  url       = {https://dl.acm.org/doi/10.1145/3534678.3542676}
}

@article{olivecrona2017molecular,
  title        = {Molecular de-novo design through deep reinforcement learning},
  author       = {Olivecrona, Marcus and Blaschke, Thomas and Engkvist, Ola and Chen, Hongming},
  journal      = {Journal of Cheminformatics},
  year         = {2017},
  volume       = {9},
  pages        = {48},
  doi          = {10.1186/s13321-017-0235-x},
  url          = {https://link.springer.com/article/10.1186/s13321-017-0235-x}
}

@inproceedings{jin2020multiobjective,
  title        = {Multi-Objective Molecule Generation using Interpretable Substructures},
  author       = {Jin, Wengong and Barzilay, Regina and Jaakkola, Tommi},
  booktitle    = {Proceedings of the 37th International Conference on Machine Learning},
  series       = {Proceedings of Machine Learning Research},
  volume       = {119},
  pages        = {4849--4859},
  year         = {2020},
  publisher    = {PMLR},
  url          = {https://proceedings.mlr.press/v119/jin20b.html}
}

@inproceedings{ijcai2024p666,
  title        = {Dynamic Many-Objective Molecular Optimization: Unfolding Complexity with Objective Decomposition and Progressive Optimization},
  author       = {Shin, Dong-Hee and Son, Young-Han and Lee, Deok-Joong and Han, Ji-Wung and Kam, Tae-Eui},
  booktitle    = {Proceedings of the Thirty-Third International Joint Conference on Artificial Intelligence, {IJCAI-24}},
  publisher    = {International Joint Conferences on Artificial Intelligence Organization},
  editor       = {Larson, Kate},
  pages        = {6026--6034},
  year         = {2024},
  month        = {8},
  note         = {Main Track},
  doi          = {10.24963/ijcai.2024/666},
  url          = {https://doi.org/10.24963/ijcai.2024/666}
}

@article{guo2024augmented,
  title        = {Augmented Memory: Sample-Efficient Generative Molecular Design with Reinforcement Learning},
  author       = {Guo, Jeff and Schwaller, Philippe},
  journal      = {JACS Au},
  year         = {2024},
  volume       = {4},
  number       = {6},
  pages        = {2160--2172},
  doi          = {10.1021/jacsau.4c00066},
  url          = {https://pubmed.ncbi.nlm.nih.gov/38938817/}
}

@misc{kim2024geneticgflownets,
  title        = {Genetic-guided {GFlowNets} for Sample Efficient Molecular Optimization},
  author       = {Kim, Hyeonah and Kim, Minsu and Choi, Sanghyeok and Park, Jinkyoo},
  year         = {2024},
  eprint       = {2402.05961},
  archivePrefix= {arXiv},
  primaryClass = {q-bio.BM},
  doi          = {10.48550/arXiv.2402.05961},
  note         = {NeurIPS 2024},
  url          = {https://arxiv.org/abs/2402.05961}
}

@inproceedings{gao2022sampleefficiency,
  title        = {Sample Efficiency Matters: A Benchmark for Practical Molecular Optimization},
  author       = {Gao, Wenhao and Fu, Tianfan and Sun, Jimeng and Coley, Connor W.},
  booktitle    = {Advances in Neural Information Processing Systems},
  year         = {2022},
  note         = {Datasets and Benchmarks Track},
  url          = {https://proceedings.neurips.cc/paper_files/paper/2022/hash/8644353f7d307baaf29bc1e56fe8e0ec-Abstract-Datasets_and_Benchmarks.html}
}

@InProceedings{pmlr-v80-jin18a,
  title     = {Junction Tree Variational Autoencoder for Molecular Graph Generation},
  author    = {Jin, Wengong and Barzilay, Regina and Jaakkola, Tommi},
  booktitle = {Proceedings of the 35th International Conference on Machine Learning},
  pages     = {2323--2332},
  year      = {2018},
  editor    = {Dy, Jennifer and Krause, Andreas},
  volume    = {80},
  series    = {Proceedings of Machine Learning Research},
  month     = {10--15 Jul},
  publisher = {PMLR},
  url       = {https://proceedings.mlr.press/v80/jin18a.html}
}

@inproceedings{jin2019graph2graph,
  title     = {Learning Multimodal Graph-to-Graph Translation for Molecule Optimization},
  author    = {Jin, Wengong and Yang, Kevin and Barzilay, Regina and Jaakkola, Tommi},
  booktitle = {International Conference on Learning Representations (ICLR)},
  year      = {2019},
  url       = {https://openreview.net/forum?id=B1xJAsA5F7}
}

@inproceedings{fu2022dst,
  title     = {Differentiable Scaffolding Tree for Molecule Optimization},
  author    = {Fu, Tianfan and Gao, Wenhao and Xiao, Cao and Yasonik, Jacob and Coley, Connor W. and Sun, Jimeng},
  booktitle = {International Conference on Learning Representations (ICLR)},
  year      = {2022},
  url       = {https://openreview.net/forum?id=w_drCosT76}
}

@InProceedings{pmlr-v202-lee23f,
  title     = {Exploring Chemical Space with Score-based Out-of-distribution Generation},
  author    = {Lee, Seul and Jo, Jaehyeong and Hwang, Sung Ju},
  booktitle = {Proceedings of the 40th International Conference on Machine Learning},
  pages     = {18872--18892},
  year      = {2023},
  editor    = {Krause, Andreas and Brunskill, Emma and Cho, Kyunghyun and Engelhardt, Barbara and Sabato, Sivan and Scarlett, Jonathan},
  volume    = {202},
  series    = {Proceedings of Machine Learning Research},
  month     = {23--29 Jul},
  publisher = {PMLR},
  url       = {https://proceedings.mlr.press/v202/lee23f.html}
}

@article{bagal2022molgpt,
  title     = {MolGPT: Molecular Generation Using a Transformer-Decoder Model},
  author    = {Bagal, Viraj and Aggarwal, Rishal and Vinod, P. K. and Priyakumar, U. Deva},
  journal   = {Journal of Chemical Information and Modeling},
  year      = {2022},
  volume    = {62},
  number    = {9},
  pages     = {2064--2076},
  doi       = {10.1021/acs.jcim.1c00600}
}

@inproceedings{fang2024molgen,
  title     = {Domain-Agnostic Molecular Generation with Chemical Feedback},
  author    = {Fang, Yin and Zhang, Ningyu and Chen, Zhuo and Guo, Lingbing and Fan, Xiaohui and Chen, Huajun},
  booktitle = {International Conference on Learning Representations (ICLR)},
  year      = {2024},
  url       = {https://openreview.net/forum?id=9rPyHyjfwP}
}

@article{abeer2024molso,
  title   = {Multi-objective latent space optimization of generative molecular design models},
  author  = {Abeer, A. N. M. Nafiz and Urban, Nathan M. and Weil, M. Ryan and Alexander, Francis J. and Yoon, Byung-Jun},
  journal = {Patterns},
  year    = {2024},
  volume  = {5},
  number  = {10},
  pages   = {101042},
  doi     = {10.1016/j.patter.2024.101042},
  url     = {https://www.sciencedirect.com/science/article/pii/S2666389924001843}
}

@article{ai4science2023impact,
  title         = {The Impact of Large Language Models on Scientific Discovery: A Preliminary Study using GPT-4},
  author        = {{Microsoft Research AI4Science} and {Microsoft Azure Quantum}},
  journal       = {arXiv preprint arXiv:2311.07361},
  year          = {2023},
  url           = {https://arxiv.org/abs/2311.07361}
}

@article{wu2024ecsurvey,
  title         = {Evolutionary Computation in the Era of Large Language Model: Survey and Roadmap},
  author        = {Wu, Xingyu and Wu, Sheng-hao and Wu, Jibin and Feng, Liang and Tan, Kay Chen},
  journal       = {arXiv preprint arXiv:2401.10034},
  year          = {2024},
  url           = {https://arxiv.org/abs/2401.10034}
}

@inproceedings{brown2020gpt3,
  title         = {Language Models are Few-Shot Learners},
  author        = {Brown, Tom B. and Mann, Benjamin and Ryder, Nick and Subbiah, Melanie and Kaplan, Jared D. and Dhariwal, Prafulla and Neelakantan, Arvind and Shyam, Pranav and Sastry, Girish and Askell, Amanda and others},
  booktitle     = {Advances in Neural Information Processing Systems (NeurIPS)},
  volume        = {33},
  year          = {2020},
  url           = {https://arxiv.org/abs/2005.14165}
}

@misc{yang2024opro,
  title         = {Large Language Models as Optimizers},
  author        = {Yang, Chengrun and Wang, Xuezhi and Lu, Yifeng and Liu, Hanxiao and Le, Quoc V. and Zhou, Denny and Chen, Xinyun},
  year          = {2024},
  note          = {arXiv:2309.03409 (last revised 15 Apr 2024); ICLR 2024},
  eprint        = {2309.03409},
  archivePrefix = {arXiv},
  primaryClass  = {cs.LG},
  doi           = {10.48550/arXiv.2309.03409},
  url           = {https://arxiv.org/abs/2309.03409}
}

@misc{liu2024lmea,
  title         = {Large Language Models as Evolutionary Optimizers},
  author        = {Liu, Shengcai and Chen, Caishun and Qu, Xinghua and Tang, Ke and Ong, Yew-Soon},
  year          = {2024},
  note          = {arXiv:2310.19046 (last revised 26 Apr 2024); accepted by CEC 2024},
  eprint        = {2310.19046},
  archivePrefix = {arXiv},
  primaryClass  = {cs.NE},
  doi           = {10.48550/arXiv.2310.19046},
  url           = {https://arxiv.org/abs/2310.19046}
}

@misc{wang2024llm_cmo,
  title         = {Large Language Model-Aided Evolutionary Search for Constrained Multiobjective Optimization},
  author        = {Wang, Zeyi and Liu, Songbai and Chen, Jianyong and Tan, Kay Chen},
  year          = {2024},
  eprint        = {2405.05767},
  archivePrefix = {arXiv},
  primaryClass  = {cs.NE},
  doi           = {10.48550/arXiv.2405.05767},
  note          = {International Conference on Intelligent Computing (ICIC) 2024},
  url           = {https://arxiv.org/abs/2405.05767}
}

@misc{liu2023ael,
  title         = {Algorithm Evolution Using Large Language Model},
  author        = {Liu, Fei and Tong, Xialiang and Yuan, Mingxuan and Zhang, Qingfu},
  year          = {2023},
  eprint        = {2311.15249},
  archivePrefix = {arXiv},
  primaryClass  = {cs.NE},
  doi           = {10.48550/arXiv.2311.15249},
  url           = {https://arxiv.org/abs/2311.15249}
}

@misc{liu2024eoh,
  title         = {Evolution of Heuristics: Towards Efficient Automatic Algorithm Design Using Large Language Model},
  author        = {Liu, Fei and Tong, Xialiang and Yuan, Mingxuan and Lin, Xi and Luo, Fu and Wang, Zhenkun and Lu, Zhichao and Zhang, Qingfu},
  year          = {2024},
  eprint        = {2401.02051},
  archivePrefix = {arXiv},
  primaryClass  = {cs.NE},
  doi           = {10.48550/arXiv.2401.02051},
  note          = {ICML 2024},
  url           = {https://arxiv.org/abs/2401.02051}
}

@misc{liu2024llm_moea,
  title         = {Large Language Model Aided Multi-objective Evolutionary Algorithm: a Low-cost Adaptive Approach},
  author        = {Liu, Wanyi and Chen, Long and Tang, Zhenzhou},
  year          = {2024},
  eprint        = {2410.02301},
  archivePrefix = {arXiv},
  primaryClass  = {cs.NE},
  doi           = {10.48550/arXiv.2410.02301},
  url           = {https://arxiv.org/abs/2410.02301}
}

@misc{huang2024blackbox_llm,
  title         = {Exploring the True Potential: Evaluating the Black-box Optimization Capability of Large Language Models},
  author        = {Huang, Beichen and Wu, Xingyu and Zhou, Yu and Wu, Jibin and Feng, Liang and Cheng, Ran and Tan, Kay Chen},
  year          = {2024},
  eprint        = {2404.06290},
  archivePrefix = {arXiv},
  primaryClass  = {cs.AI},
  doi           = {10.48550/arXiv.2404.06290},
  url           = {https://arxiv.org/abs/2404.06290}
}

@misc{brahmachary2024leo,
  title         = {Large Language Model-Based Evolutionary Optimizer: Reasoning with elitism},
  author        = {Brahmachary, Shuvayan and Joshi, Subodh M. and Panda, Aniruddha and Koneripalli, Kaushik and Sagotra, Arun Kumar and Patel, Harshil and Sharma, Ankush and Jagtap, Ameya D. and Kalyanaraman, Kaushic},
  year          = {2024},
  eprint        = {2403.02054},
  archivePrefix = {arXiv},
  primaryClass  = {cs.AI},
  doi           = {10.48550/arXiv.2403.02054},
  url           = {https://arxiv.org/abs/2403.02054}
}

@article{zeng2022deep,
  title={A deep-learning system bridging molecule structure and biomedical text with comprehension comparable to human professionals},
  author={Zeng, Zheni and Yao, Yuan and Liu, Zhiyuan and Sun, Maosong},
  journal={Nature communications},
  volume={13},
  number={1},
  pages={862},
  year={2022},
  publisher={Nature Publishing Group UK London}
}

@article{zeng2024chatmol,
  title={ChatMol: interactive molecular discovery with natural language},
  author={Zeng, Zheni and Yin, Bangchen and Wang, Shipeng and Liu, Jiarui and Yang, Cheng and Yao, Haishen and Sun, Xingzhi and Sun, Maosong and Xie, Guotong and Liu, Zhiyuan},
  journal={Bioinformatics},
  volume={40},
  number={9},
  pages={btae534},
  year={2024},
  publisher={Oxford University Press}
}

@article{zhang2025computational_toxicology_admet,
  title   = {Computational toxicology in drug discovery: applications of artificial intelligence in {ADMET} and toxicity prediction},
  author  = {Zhang, Jiangyan and Li, Haolin and Zhang, Yuncong and Huang, Junyang and Ren, Liping and Zhang, Chuantao and Zou, Quan and Zhang, Yang},
  journal = {Briefings in Bioinformatics},
  volume  = {26},
  number  = {5},
  pages   = {bbaf533},
  year    = {2025},
  month   = sep,
  doi     = {10.1093/bib/bbaf533}
}

@article{meng2011docking,
  title   = {Molecular Docking: A Powerful Approach for Structure-Based Drug Discovery},
  author  = {Meng, Xuan-Yu and Zhang, Hong-Xing and Mezei, Mihaly and Cui, Meng},
  journal = {Current Computer-Aided Drug Design},
  volume  = {7},
  number  = {2},
  pages   = {146--157},
  year    = {2011},
  doi     = {10.2174/157340911795677602}
}

@article{ferreira2015docking,
  title   = {Molecular Docking and Structure-Based Drug Design Strategies},
  author  = {Ferreira, Leonardo G. and dos Santos, Ricardo N. and Oliva, Glaucius and Andricopulo, Adriano D.},
  journal = {Molecules},
  volume  = {20},
  number  = {7},
  pages   = {13384--13421},
  year    = {2015},
  doi     = {10.3390/molecules200713384}
}

@article{lu2010residencetime,
  title   = {Drug-Target Residence Time: Critical Information for Lead Optimization},
  author  = {Lu, Hao and Tonge, Peter J.},
  journal = {Current Opinion in Chemical Biology},
  volume  = {14},
  number  = {4},
  pages   = {467--474},
  year    = {2010},
  doi     = {10.1016/j.cbpa.2010.06.176}
}

@article{riniker2015etkdg,
  title   = {Better Informed Distance Geometry: Using What We Know To Improve Conformation Generation},
  author  = {Riniker, Sereina and Landrum, Gregory A.},
  journal = {Journal of Chemical Information and Modeling},
  volume  = {55},
  number  = {12},
  pages   = {2562--2574},
  year    = {2015},
  doi     = {10.1021/acs.jcim.5b00654}
}

@article{halgren1996mmff,
  title   = {Merck Molecular Force Field. I. Basis, Form, Scope, Parameterization, and Performance of {MMFF94}},
  author  = {Halgren, Thomas A.},
  journal = {Journal of Computational Chemistry},
  volume  = {17},
  number  = {5-6},
  pages   = {490--519},
  year    = {1996},
  doi     = {10.1002/(SICI)1096-987X(199604)17:5/6<490::AID-JCC1>3.0.CO;2-P}
}

@article{oboyle2011openbabel,
  title   = {Open Babel: An Open Chemical Toolbox},
  author  = {O'Boyle, Noel M. and Banck, Michael and James, Craig A. and Morley, Chris and Vandermeersch, Tim and Hutchison, Geoffrey R.},
  journal = {Journal of Cheminformatics},
  volume  = {3},
  pages   = {33},
  year    = {2011},
  doi     = {10.1186/1758-2946-3-33}
}

@article{gasteiger1980charges,
  title   = {Iterative Partial Equalization of Orbital Electronegativity---A Rapid Access to Atomic Charges},
  author  = {Gasteiger, Johann and Marsili, Mario},
  journal = {Tetrahedron},
  volume  = {36},
  number  = {22},
  pages   = {3219--3228},
  year    = {1980},
  doi     = {10.1016/0040-4020(80)80168-2}
}

@article{trott2010vina,
  title   = {AutoDock Vina: Improving the Speed and Accuracy of Docking with a New Scoring Function, Efficient Optimization, and Multithreading},
  author  = {Trott, Oleg and Olson, Arthur J.},
  journal = {Journal of Computational Chemistry},
  volume  = {31},
  number  = {2},
  pages   = {455--461},
  year    = {2010},
  doi     = {10.1002/jcc.21334}
}

@article{guan2019admet,
  title        = {ADMET-score--a comprehensive scoring function for evaluation of chemical drug-likeness},
  author       = {Guan, L and Yang, H and Cai, Y and Sun, L and Di, P and Li, W and Liu, G and Tang, Y},
  journal      = {MedChemComm},
  year         = {2019},
  volume       = {10},
  pages        = {148--157},
  doi          = {10.1039/C8MD00472B},
  publisher    = {Royal Society of Chemistry},
}

@online{openai_gpt4omini_docs,
  author = {{OpenAI}},
  title  = {GPT-4o mini (Model Documentation)},
  year   = {2024},
  url    = {https://platform.openai.com/docs/models/gpt-4o-mini},
  note   = {Accessed: 2026-01-28}
}

@article{khadem2025cad,
  title   = {In-silico identification of novel Cis-aconitate decarboxylase inhibitors as potential anti-inflammatory agents using molecular docking and dynamics},
  author  = {Khadem, Mohammad Darvish and Pirmoradi, Saeed and Tabandeh, Mohammad Reza and Zarezade, Vahid and Monjezi, Zohre and others},
  journal = {Scientific Reports},
  volume  = {15},
  year    = {2025},
  pages   = {42412},
  doi     = {10.1038/s41598-025-26473-4},
  url     = {https://www.nature.com/articles/s41598-025-26473-4}
}

@article{zhang2024deepgpcr,
  title   = {Revolutionizing GPCR--ligand predictions: DeepGPCR with experimental validation for high-precision drug discovery},
  author  = {Zhang, Haiping and Fan, Hongjie and Wang, Jixia and Hou, Tao and Saravanan, Konda Mani and Xia, Wei and Kan, Hei Wun and Li, Junxin and Zhang, John Z. H and Liang, Xinmiao and Chen, Yang and others},
  journal = {Briefings in Bioinformatics},
  volume  = {25},
  number  = {4},
  year    = {2024},
  pages   = {bbae281},
  doi     = {10.1093/bib/bbae281},
  url     = {https://academic.oup.com/bib/article/25/4/bbae281/7691386}
}

@article{langevin2022failure,
  title   = {Explaining and avoiding failure modes in goal-directed generation of small molecules},
  author  = {Langevin, Maxime and Vuilleumier, Rodolphe and Bianciotto, Marc},
  journal = {Journal of Cheminformatics},
  volume  = {14},
  number  = {1},
  pages   = {20},
  year    = {2022},
  doi     = {10.1186/s13321-022-00601-y},
  url     = {https://link.springer.com/article/10.1186/s13321-022-00601-y}
}

@article{yoshizawa2025rewardhacking,
  title   = {A data-driven generative strategy to avoid reward hacking in multi-objective molecular design},
  author  = {Yoshizawa, T. and others},
  journal = {Nature Communications},
  volume  = {16},
  year    = {2025},
  pages   = {2409},
  doi     = {10.1038/s41467-025-57582-3},
  url     = {https://www.nature.com/articles/s41467-025-57582-3}
}

@article{narayanan2025ether0,
  title         = {Training a Scientific Reasoning Model for Chemistry},
  author        = {Narayanan, Siddharth M. and Braza, James D. and Griffiths, Ryan-Rhys and Bou, Albert and Wellawatte, Geemi and Caldas Ramos, Mayk and Mitchener, Ludovico and Rodriques, Samuel G. and White, Andrew D.},
  year          = {2025},
  journal       = {arXiv preprint arXiv:2506.17238},
  eprint        = {2506.17238},
  archivePrefix = {arXiv},
  primaryClass  = {cs.LG},
  doi           = {10.48550/arXiv.2506.17238},
  url           = {https://arxiv.org/abs/2506.17238},
  note          = {Introduces the chemistry reasoning model \textit{ether0}.}
}

@article{zhao2024chemdfm,
  title         = {Developing {ChemDFM} as a Large Language Foundation Model for Chemistry},
  author        = {Zhao, Zihan and Ma, Da and Chen, Lu and Sun, Liangtai and Li, Zihao and Xia, Yi and Chen, Bo and Xu, Hongshen and Zhu, Zichen and Zhu, Su and Fan, Shuai and Shen, Guodong and Yu, Kai and Chen, Xin},
  year          = {2024},
  journal       = {arXiv preprint arXiv:2401.14818},
  eprint        = {2401.14818},
  archivePrefix = {arXiv},
  primaryClass  = {cs.CL},
  doi           = {10.48550/arXiv.2401.14818},
  url           = {https://arxiv.org/abs/2401.14818},
  note          = {Journal reference: Cell Rep. Phys. Sci. 6 (2025) 102523.}
}

@article{zhang2024chemllm,
  title         = {{ChemLLM}: A Chemical Large Language Model},
  author        = {Zhang, Di and Liu, Wei and Tan, Qian and Chen, Jingdan and Yan, Hang and Yan, Yuliang and Li, Jiatong and Huang, Weiran and Yue, Xiangyu and Ouyang, Wanli and Zhou, Dongzhan and Zhang, Shufei and Su, Mao and Zhong, Han-Sen and Li, Yuqiang},
  year          = {2024},
  journal       = {arXiv preprint arXiv:2402.06852},
  eprint        = {2402.06852},
  archivePrefix = {arXiv},
  primaryClass  = {cs.AI},
  doi           = {10.48550/arXiv.2402.06852},
  url           = {https://arxiv.org/abs/2402.06852}
}

@article{wu2018moleculenet,
  title   = {MoleculeNet: a benchmark for molecular machine learning},
  author  = {Wu, Zhenqin and Ramsundar, Bharath and Feinberg, Evan N. and Gomes, Joseph and Geniesse, Caleb and Pappu, Aneesh S. and Leswing, Karl and Pande, Vijay},
  journal = {Chemical Science},
  volume  = {9},
  number  = {2},
  pages   = {513--530},
  year    = {2018},
  doi     = {10.1039/C7SC02664A}
}

@article{shao2024deepseekmath,
  title={Deepseekmath: Pushing the limits of mathematical reasoning in open language models},
  author={Shao, Zhihong and Wang, Peiyi and Zhu, Qihao and Xu, Runxin and Song, Junxiao and Bi, Xiao and Zhang, Haowei and Zhang, Mingchuan and Li, YK and others},
  journal={arXiv preprint arXiv:2402.03300},
  year={2024}
}

@inproceedings{ding-etal-2023-enhancing,
  title     = {Enhancing Chat Language Models by Scaling High-quality Instructional Conversations},
  author    = {Ding, Ning and Chen, Yulin and Xu, Bokai and Qin, Yujia and Hu, Shengding and Liu, Zhiyuan and Sun, Maosong and Zhou, Bowen},
  booktitle = {Proceedings of the 2023 Conference on Empirical Methods in Natural Language Processing},
  year      = {2023},
  pages     = {3029--3051},
  doi       = {10.18653/v1/2023.emnlp-main.183}
}

@misc{chemqa_internal,
  title  = {ChemicalQA},
  author = {{This work}},
  note   = {Internal chemical QA corpus used for domain-adaptive continued pretraining.}
}

@misc{cpt_internal,
  title  = {General-domain CPT text},
  author = {{This work}},
  note   = {Internal general-domain text corpus used for continued pretraining.}
}

@misc{ran2025exllm,
  title         = {ExLLM: Experience-Enhanced LLM Optimization for Molecular Design and Beyond},
  author        = {Ran, Nian and Wang, Yue and Zhang, Xiaoyuan and Li, Zhongzheng and Ran, Qingsong and Li, Wenhao and Allmendinger, Richard},
  year          = {2025},
  eprint        = {2502.12845},
  archivePrefix = {arXiv},
  primaryClass  = {cs.LG},
  doi           = {10.48550/arXiv.2502.12845},
  url           = {https://arxiv.org/abs/2502.12845}
}

@article{ChangYe2024LDMol,
  title        = {LDMol: A Text-to-Molecule Diffusion Model with Structurally Informative Latent Space Surpasses AR Models},
  author       = {Jinho Chang and Jong Chul Ye},
  journal      = {arXiv preprint arXiv:2405.17829},
  year         = {2024},
  doi          = {10.48550/arXiv.2405.17829},
  url          = {https://arxiv.org/abs/2405.17829}
}

@inproceedings{Zhu2023SampleEfficientMO,
  title        = {Sample-efficient Multi-objective Molecular Optimization with GFlowNets},
  author       = {Yiheng Zhu and Jialu Wu and Chaowen Hu and Jiahuan Yan and Chang-Yu Hsieh and Tingjun Hou and Jian Wu},
  booktitle    = {Proceedings of the Thirty-Seventh Conference on Neural Information Processing Systems (NeurIPS 2023)},
  year         = {2023},
  url          = {https://proceedings.neurips.cc/paper_files/paper/2023/hash/fbc9981dd6316378aee7fd5975250f21-Abstract-Conference.html},
}
\bibliographystyle{rsc} 
\section*{Supplementary Information}
\subsubsection*{Template S1: Logical coherence evaluation for molecular optimization rationales}
\vspace{-0.5em}
\begin{lstlisting}
You are an expert in logical analysis. Please evaluate the logical coherence of the following molecular optimization rationale.
Please assess the following key aspects:
1) Alignment between problem identification and proposed solution (score 0-1, where 1 = perfect alignment)
- Are the identified problems clearly stated?
- Does the proposed solution directly address these problems?
- Is there any mismatch such as "Problem A is raised, but the solution targets Problem B"?
2) Alignment between the proposed solution and the SMILES edits (score 0-1, where 1 = perfect alignment)
- Is the modification strategy reflected in the SMILES?
- Do the actual edits match the described strategy?
- Is there any mismatch such as "Claims to add a substituent, but actually removes one"?
3) Completeness of the overall logical chain (score 0-1, where 1 = complete)
Is the chain "problem -> strategy -> modification -> expected effect" coherent and continuous?
4) Accuracy of causal reasoning (score 0-1, where 1 = accurate)
Is the "because X, therefore Y" reasoning sound? Is there any reversal of cause and effect?
\end{lstlisting}

\subsection*{Template S2: Prompting template for rationale generation}
\vspace{-0.5em}
\begin{lstlisting}
Please provide a detailed ADMET-focused rationale (5-8 sentences) about the key structural edits and why they improve ADMET properties. Include specific analysis of:
1. Key structural changes made
2. How these changes address the identified ADMET issues
3. Specific ADMET property improvements (absorption, metabolism, toxicity, synthesis)
4. Molecular mechanisms behind the improvements
5. Overall drug-likeness enhancement
\end{lstlisting}

\subsection*{Template S3: Prompt templates and specifications for data generation and formatting}
\vspace{-0.5em}
\begin{lstlisting}
You are a medicinal chemistry assistant. Given an original molecule (SMILES) and its ADMET profile, design a single optimized molecule that best improves the stated ADMET liabilities while preserving potency.
Write naturally: first give a brief rationale (6-8 sentences) explaining the key modifications and why they help. Then provide exactly one SMILES for your best design in a code block.
\end{lstlisting}

\end{document}